\documentclass[preprint]{elsarticle}

\usepackage{hyperref}
\usepackage{times}
\usepackage{epsfig}
\usepackage{graphicx}
\usepackage{amsmath}
\usepackage{amssymb}
\usepackage{algorithm} 
\usepackage{algpseudocode}

\usepackage{xcolor}
\usepackage{amsthm}
\usepackage{subcaption}
\usepackage{graphicx}

\usepackage{cleveref}

\theoremstyle{definition}
\newtheorem{definition}{Definition}

\theoremstyle{definition}
\newtheorem{hypothesis}{Hypothesis}

\newcommand{\softmax}{softmax}
\newcommand{\ce}{CrossEntropy}

\newcommand{\res}{res}

\journal{Neural Networks}

\begin{document}

\begin{frontmatter}

\title{Auto-tuning of Deep Neural Networks by \\ Conflicting Layer Removal}

\author{David Peer\footnote{david.peer@outlook.com}, Sebastian Stabinger, Antonio Rodríguez-Sánchez}
\address{Universität Innsbruck, Technikerstrasse 21a, 6020 Innsbruck, Austria}

\begin{abstract}
Designing neural network architectures is a challenging task and knowing which specific layers of a model must be adapted to improve the performance is almost a mystery. In this paper, we introduce a novel methodology to identify layers that decrease the test accuracy of trained models. Conflicting layers are detected as early as the beginning of training. In the worst-case scenario, we prove that such a layer could lead to a network that cannot be trained at all. A theoretical analysis is provided on what is the origin of those layers that result in a lower overall network performance, which is complemented by our extensive empirical evaluation. More precisely, we identified those layers that worsen the performance because they would produce what we name \textit{conflicting training bundles}. We will show that around $60\%$ of the layers of trained residual networks can be completely removed from the architecture with no significant increase in the test-error. We will further present a novel neural-architecture-search (NAS) algorithm that identifies \textit{conflicting} layers at the beginning of the training. Architectures found by our auto-tuning algorithm achieve competitive accuracy values when compared against more complex state-of-the-art architectures, while drastically reducing memory consumption and inference time for different computer vision tasks. The source code is available on  \url{https://github.com/peerdavid/conflicting-bundles}
\end{abstract}


\end{frontmatter}


\section{Introduction}\label{sec:introduction}

\begin{figure}[t]
  \centering
  \captionsetup{justification=justified}
  \includegraphics[width=.65\linewidth]{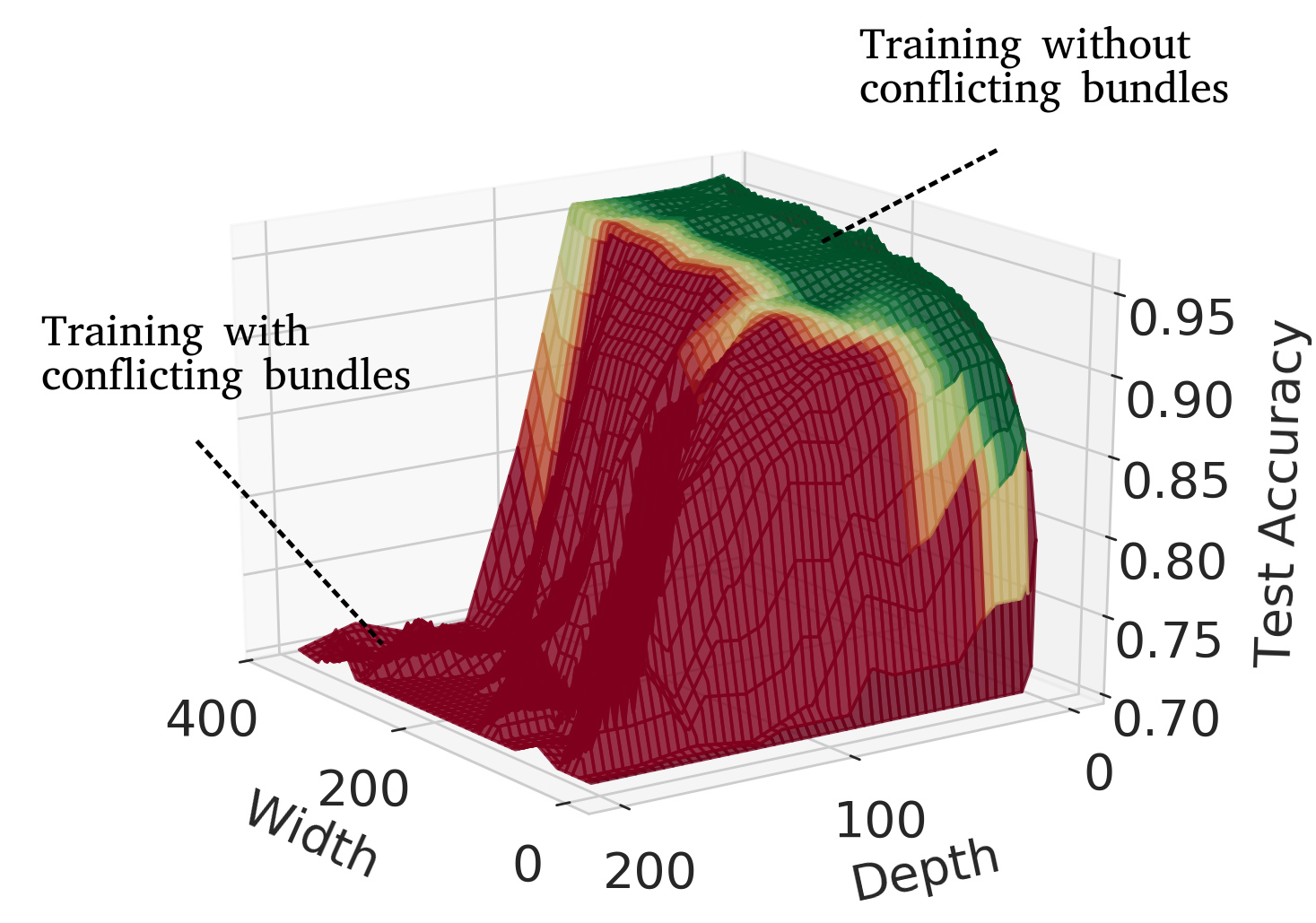}
  \caption{Test accuracy of 176 neural networks at different depths and widths trained on MNIST together with the boundary (yellow) that separates networks that induce conflicting training bundles (red) from those free of conflicting bundles (green).}\label{fig:experiment_fnn_first_conflicting_layer}
\end{figure}

The training of deep neural networks is a complex and challenging task \cite{unsupervised_pretraining}, one that can be achieved by better initialization strategies \cite{glorot_initializer, he_initializer}, activation functions \cite{relu, mish}, regularization methods \cite{batch_normalization, dropout} and network architectures \cite{residual_neural_networks, gammacapsules, highway_networks, efficientnet}. It is known that the expressivity of the networks grows exponentially with depth \cite{capsnet_limitations, expressive_power}, but designing architectures that achieve high accuracy on a given task is challenging and can not only be achieved through the creation of deeper networks \cite{efficientnet}. \Cref{fig:experiment_fnn_first_conflicting_layer} shows the accuracy of $176$ different fully connected neural networks trained on MNIST at different depths and widths. We can observe that the optimal architecture w.r.t test accuracy that is trainable depends not only on depth, but also on a good balance between width and depth. Searching through different architectures in this manner in order to find a good configuration is not reasonable for real-world applications as too many possible configurations exist and the time required to train and evaluate each alternative is excessively long \cite{how_many_layers_and_nodes}. We will show theoretically and experimentally that  \emph{conflicting bundles} worsen the performance of neural network models. Models trained with conflicting bundles are shown in red in \cref{fig:experiment_fnn_first_conflicting_layer}, while the green color represents the models that are trained without conflicting bundles. The difference in accuracy values is quite noticeable. We will show that we are able to locate the layers that induce conflicting bundles already at the beginning of training, making our method an efficient strategy for dismissing underperforming architectures. The analysis and study of \emph{conflicting bundles} we present in this paper will: 

\begin{itemize}
    \item Explain from a new theoretical perspective why a balance between width and depth is important, showing why residual connections help to train very deep networks.
    \item Help researchers to create more efficient state-of-the-art network architectures, which we will demonstrate by pruning as much as $60\%$ of the layers of an already trained residual network without decreasing the accuracy significantly.
    \item Set the grounds for additional research pathways in Auto-ML
\end{itemize}

In this work, we further study the \emph{conflicting training bundle problem}, as introduced in our previous work \cite{conflicting_bundles_wacv}, providing a deeper theoretical analysis through a detailed derivation problem breakdown, which allows us to present an innovative methodology to prune residual networks. We will then provide additional experimental evaluation consisting of a lesion experiment, including a comparison to the work from \citet{residual_behave_shallow}, which is later complemented with further evaluation on other residual connections than the identity mapping.

\subsection{The conflicting training bundle problem} \label{sec:bundle_problem}

The output of a neural network is calculated by successively propagating inputs forward through all hidden layers. All output values of a layer are represented with finite resolution (e.g. 32 bit floating-point values). Consequently, two outputs that are only slightly different (less than the minimum resolution of the used floating-point representation), can become equal.

We have found that during training, weights were adjusted in wrong directions - leading to a worsened overall performance of the model - if any hidden layer produces the same output vector for two input examples with different labels. We call two samples \emph{bundled} if the same output w.r.t. the floating-point resolution is produced for both inputs after passing through some hidden layer of the network. If both samples are labeled differently, we name them \emph{conflicting}. We are able to determine the layer where the samples are bundled - called from now on, the \emph{conflicting layer} - and therefore, non-optimal architectures can either manually or automatically be adapted to be more accurate and efficient as we will show in this paper.

\subsection{Outline}
In \cref{sec:theory}, conflicting bundles are formally defined and their impact during training is analyzed. We show theoretically that the accuracy of a model is bound to decrease if conflicting bundles appear during training. In the worst-case scenario, the network will just not learn from data. In that same section, we introduce a novel metric to quantify and detect the precise layer that produces conflicting training bundles. In the experimental \cref{sec:experimental}, we will evaluate a broad range of different types of networks. Under controlled settings, we produce conflicting training bundles and study their effects per training epoch and layer. For completeness, fully connected networks, VGG nets, and ResNets are evaluated on different datasets. A detailed theoretical and empirical analysis is provided for residual connections which finally leads to a novel pruning method for ResNets that can be used to prune up to $60\%$ of the layers from already trained ResNets 
without loss of accuracy. A novel NAS (neural-architecture-search) algorithm to tackle conflicting training bundles already at the beginning of the training is also introduced and compared against manually designed network architectures. A discussion and inspiration for future research are given in \cref{sec:discussion}. 

\subsection{Related work}\label{sec:related_work}
Solving different optimization problems have been the subject of extensive study. \citet{efficient_backprop} has shown that a careful initialization of the weights of neural networks has a significant effect on the training process. Methods to initialize weights avoid vanishing information during forward-propagation and also avoid vanishing- or exploding gradients during backward propagation \cite{glorot_initializer, he_initializer, unsupervised_pretraining, data_dependent_init, need_good_init}. Very deep networks are difficult to optimize, even when variance-preserving initialization methods are used \cite{highway_networks}. This problem can be overcome by highway networks from  \citet{highway_networks} that allow for the unimpeded information flow across several layers or residual learning as presented by  \citet{residual_neural_networks}. Historical developments and optimization problems that can occur during the training of neural networks are extensively described and summarized by the seminal survey of \citet{deep_learning_in_nn_overview} and the book by \citet{deep_learning_book}.

To the best of our knowledge, we are the first to precisely locate and quantify the conflicting training bundles problem. The problem we have identified is related to, but different from, the Vanishing information problem \cite{vanishing_information}. They describe that information of original input patterns is lost in higher layers by going through multiple layer transformations and compressions. \citet{vanishing_information} also showed that convolutional neural networks do not suffer from this problem, probably due to the high dimensionality of hidden features. In the case of conflicting bundles, two different inputs are represented as equal when passing through a layer and, in contrast to the vanishing information problem, conflicting bundles also appear in convolutional networks. In the same domain of problems we find Shattered gradients \cite{shattered_gradients}, where the correlation between gradients in fully connected networks with $L$ layers decay exponentially with $1/2^L$. The correlation decreases only with $1 / \sqrt{L}$ if residual connections are used. Due to this fact, residual networks seem to be easier to train than fully connected networks. Shattered gradients are independent of the network's width (theorem 1 of \cite{shattered_gradients}). On the other hand, conflicting training bundles depend on both, depth and width as we will show later in this paper (\cref{sec:experiment_fnn}). Furthermore, conflicting bundles occur in very shallow networks, which rules out the vanishing- or exploding gradients problem \cite{lstm}. Finally, different to the studies in the same domain of problems, where the focus is on mainly analyzing gradients during back-propagation, the analysis to discover conflicting bundles is performed during forward-propagation, which allows us to precisely detect the layer that introduces the problem.

\section{Theoretical analysis}\label{sec:theory}

We will put our focus - for simplicity - on classification problems, although this analysis can be extended to regression problems. Let's consider a training set $\mathcal{S} \in \mathcal{X} \times \mathcal{Y}$ that contains objects from a specific domain $\mathcal{X}$ along with its labels $\mathcal{Y}$. Labels $y \in \mathcal{Y}$ are one-hot encoded and the dimensionality of the labels is $N_c$. If the input to a layer of a neural network is of dimension $m$ and the output of dimension $n$, we use weight matrices $W \in \mathbb{R}^{n \times m}$ and bias terms $b \in \mathbb{R}^{n \times 1}$ to calculate the output. The output of a specific layer $l \in \{1, ..., L\}$ for a network with $L$ layers is a vector $a^{(l+1)}(x_i) = f \left({W}^{(l)} {a}_i^{(l)} + {b}^{(l)} \right)$ for some input $x_i \in \mathcal{X}$ and nonlinearity $f$. For simplicity, we will write $a^{(l+1)}_i$ instead of $a^{(l+1)}(x_i)$, where $a^1_i = x_i$. Neural networks are trained using gradient descent with batches $\mathcal{B} \subseteq \mathcal{X}$. We assume without loss of generality that batches are uniformly distributed w.r.t. the class labels. The loss of the neural network for input $x_i$ is calculated with $J_i = \ce(h_i, y_i)$, where ${h_i} = \softmax({W}^{(L)} {a}_i^{(L)} + {b}^{(L)})$.
The gradient is calculated with $\frac{\partial J_i}{\partial {W}^{(L)}} = ({h_i} - {y_i}) {{a}^{(L)}}^T$, which for a mini-batch $\mathcal{B}$ would be $\frac{\partial J}{\partial {W}^{(L)}} = \frac{1}{|\mathcal{B}|} \sum^{|\mathcal{B}|}_{i=1} \frac{\partial J_i}{\partial {W}^{(L)}}$. 

Two different input examples $x_i$ and $x_j$ are bundled, if the same output is produced for both inputs after the non-linearity of some hidden layer $l$ is applied (the output layer is not included). This leads to the following definitions:

\begin{definition} \label{def:bundling}
    Two samples $x_i, x_j \in \mathcal{X}$
    are \emph{bundled} in layer $l$, if $a^{(l+1)}_i = a^{(l+1)}_j$ for the current configuration of learnable parameters.
\end{definition}
\begin{definition} \label{def:conflicts}
    Two samples $x_i, x_j \in \mathcal{X}$
    are \emph{conflicting}, 
    if $y_i \neq y_j$ and there exists some layer $l \in \{1, ..., L\}$ that bundles $x_i$ and $x_j$. We call layer $l$ the \emph{conflicting layer} for $x_i$ and $x_j$.
\end{definition}

We next define the concept of a \textit{bundle} and under which circumstances we call a bundle or a layer a \textit{conflicting} one:

\begin{definition} \label{def:bundle}
    A \emph{bundle} $D_i^l$ contains all samples $x \in \mathcal{B}$ that are bundled after layer $l$ and the set of all bundles $\mathcal{D}^l = \{D^l_1, D^l_2, ... D^l_k\}$ with $1 \leq k \leq |\mathcal{B}|$ ensure that $\forall i \neq j: D^l_i \cap D^l_j = \emptyset$ and $D^l_1 \cup D^l_2, ... \cup D^l_k = \mathcal{B}$.
    A bundle that contains conflicting samples  is called a \emph{conflicting bundle}.
\end{definition}

\begin{definition} \label{def:fully_conflicting_bundle}
For the special case where all samples are bundled into a single bundle i.e. $\mathcal{D}^l = \{D^l\}$, we will call $D^l_1$ a \emph{fully conflicting bundle}.
\end{definition}

\begin{figure*}[t]
\centering
\begin{subfigure}{.30\textwidth}
  \centering
  \vspace{10pt}
  \captionsetup{justification=centering}
  \includegraphics[width=.9\linewidth]{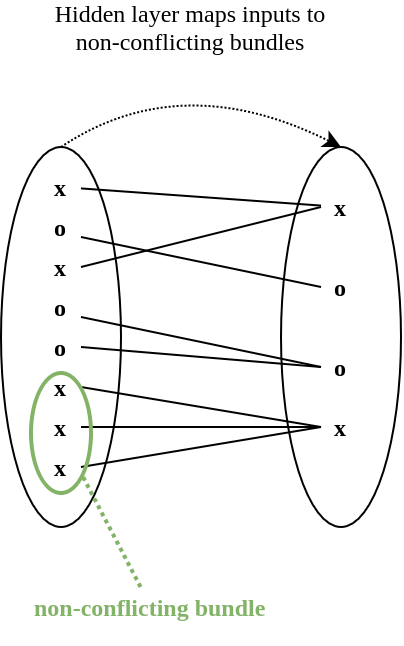}
  \caption{Layer that produces non-conflicting bundles}\label{fig:conflicting_bundle_description_a}
\end{subfigure}
\begin{subfigure}{.30\textwidth}
  \centering
  \captionsetup{justification=centering}
  \includegraphics[width=.9\linewidth]{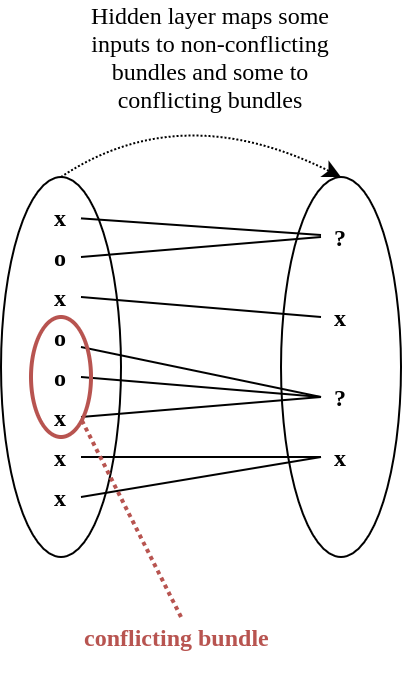}  
  \caption{Layer that produces conflicting bundles}\label{fig:conflicting_bundle_description_b}
\end{subfigure}
\begin{subfigure}{.30\textwidth}
  \centering
  \vspace{5pt}
  \captionsetup{justification=centering}
  \includegraphics[width=.9\linewidth]{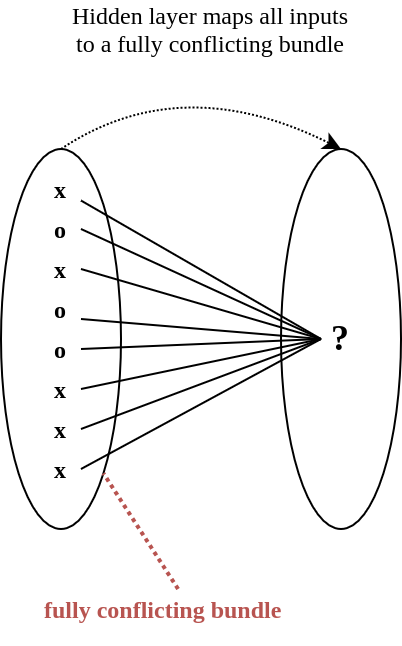}  
  \caption{Layer that produces a fully conflicting bundle}
  \label{fig:conflicting_bundle_description_c}
\end{subfigure}

\caption{Graphical illustration of non-conflicting bundles (a) conflicting bundles (b) and a fully conflicting bundle (b) that are introduced by a hidden layer for inputs of two different classes (represented as 'x' and 'o').}
\label{fig:conflicting_bundle_description}
\end{figure*}

By definition, a \emph{conflicting bundle} contains at least two samples with different labels and a \emph{fully conflicting bundle} contains all samples of a mini-batch and therefore all samples are conflicting. A graphical illustration is shown in \cref{fig:conflicting_bundle_description} to compare non-conflicting bundles, conflicting-bundles and fully-conflicting bundles. We will now make use of the definitions to derive the negative effects that arise when neural networks are trained with fully conflicting bundles or conflicting bundles in general.

\subsection{Fully conflicting bundle}\label{sec:theory_fully}
First, we will evaluate the gradient w.r.t. $W^{(L)}$ assuming that a fully conflicting bundle (\cref{def:fully_conflicting_bundle}) occurs during training:
\begin{align*}
    \frac{\partial J}{\partial {W}^{(L)}} &= \frac{1}{|\mathcal{B}|} \sum^{|\mathcal{B}|}_{i=1} \frac{\partial J_i}{\partial {W}^{(L)}} \\
    &= \frac{1}{|\mathcal{B}|} \sum^{|\mathcal{B}|}_{i=1} \left({h_i} - {y_i} \right) {{a_i}^{(L)}}^T && \text{Mini-batch gradient} \\
    &= \frac{1}{|\mathcal{B}|} \sum^{|\mathcal{B}|}_{i=1} \left({h} - {y_i} \right) {{a}^{(L)}}^T  &&\text{Fully conflicting bundle assumption} \\
    &= \frac{1}{|\mathcal{B}|} \left(|\mathcal{B}| {h} - \sum^{|\mathcal{B}|}_{i=1} {y_i} \right) {{a}^{(L)}}^T \\
    &= \frac{1}{|\mathcal{B}|} \left(|\mathcal{B}| {h} - \frac{|\mathcal{B}|}{N_c} \pmb{1} \right) {{a}^{(L)}}^T
    && \text{Uniform batches assumption}\\
    &= \left(h -  \frac{1}{N_c} \pmb{1} \right) {{a}^{(L)}}^T
\end{align*}
where $\pmb{1}$ is a vector of dimension $N_c \times 1$ with unitary elements. First of all, we can see that all $h_i$ are collapsed into a single $h$, which is the fully conflicting bundle assumption. All labels can also be collapsed into a single $\pmb{1}$ vector scaled by $|B| / N_c$ since we assumed uniformly distributed batches. The first observation is that the labels $y_i$ disappeared from the gradient, i.e. the gradient is uncorrelated of $y_i$ and therefore we conclude that the network cannot learn from data. Please note that this is different from shattered gradients \cite{shattered_gradients}, where gradients become uncorrelated from its input $x_i$ (and not $y_i$). Another observation is that the gradient becomes zero whenever all components of $h$ are equal to $1 / N_c$ and therefore we hypothesize:

\begin{hypothesis}\label{hyp:fully}
If a fully conflicting bundle occurs anywhere in the hierarchy of a neural network during training, all labels are ignored and weights are adjusted until each output neuron fires with constant value $\frac{1}{N_c}$.
\end{hypothesis}

Note that the value of each neuron becomes $\frac{1}{N_c}$ because we assumed equally distributed labels in the batches. If we relax this assumption, each neuron will fire with a constant value that represents the imbalance of the dataset. For example, if $75\%$ of the examples are of class one, the corresponding neuron will fire with a constant value of $0.75$.

\subsection{Conflicting bundles} \label{sec:theory_partially}
We now relax the assumption that a \emph{fully} conflicting bundle occurs and study the gradient for the case where some conflicting bundles occur (see \cref{def:bundle}):
\begin{align*}
    \frac{\partial J}{\partial {W}^{(L)}} &= \frac{1}{|\mathcal{B}|} \sum^{|\mathcal{B}|}_{i=1} \frac{\partial J_i}{\partial {W}^{(L)}} \\
    &= \frac{1}{|\mathcal{B}|} \sum^{|\mathcal{B}|}_{i=1} \left({h_i} - {y_i} \right) {{a_i}^{(L)}}^T & \text{Mini-batch gradient} \\
    &= \frac{1}{|\mathcal{B}|} \sum_{D \in \mathcal{D}^L} \left( |D| \ h_D -  \sum_{i=1}^{|D|} y_i \right) {{a_D}^{(L)}}^T  & \text{Conflicting bundles assumption}\\
\end{align*}
Compared to the fully conflicting bundle case, where labels disappeared, labels in this case are included in the calculation of the gradient. Unfortunately, for the same output $h_D$ different labels are grouped by $\sum_{i=1}^{|D|} y_i$. We can then think of the effect of conflicting bundles as being similar to the effect of noisy labels, for which it is well known that the model performance is worsened \cite{label_noise}. This leads us to the following hypothesis that will be confirmed in the experimental \cref{sec:experimental}:

\begin{hypothesis}\label{hyp:partially}
If conflicting bundles occur anywhere in the hierarchy of a neural network during training, the accuracy of the trained model is worsened.
\end{hypothesis}

To check the correctness of \cref{hyp:fully} and \cref{hyp:partially}, we have to be able to detect situations where conflicting bundles occur and measure the degree of conflict of a bundle. A metric to quantify conflicting bundles is introduced next.

\subsection{Conflicting bundle metric}
To measure whether two samples $x_i$ and $x_j$ are bundled at layer $l$ we must check if $a_i^{(l+1)} = a_j^{(l+1)}$ (\cref{def:bundling}). The finite resolution of floating point values must be considered. It then makes a difference whether the vectors are used during forward- or backpropagation: During backpropagation, $a_i$ vectors are scaled by the learning rate $\alpha$ and $1 / |\mathcal{B}|$ before they are subtracted from the weights. Therefore, it is possible that very small values that are different during forward propagation are bundled during backpropagation due to the finite resolution of floating-point values. To consider this situation, we approximate \cref{def:bundling} with
\begin{align} \label{eq:bundled}
    \frac{\alpha}{|\mathcal{B}|} \ ||a_i^{(l+1)} - a_j^{(l+1)}||_\infty \leq \res(W^l)
\end{align}
where $\res(W^l)$ is the smallest possible resolution that is supported by the floating-point representation w.r.t the weights $W^l$ of the GPU or CPU at hand. Note that \cref{eq:bundled} depends only on the output of the layer such that this metric can be used for any type of layers, i.e. fully connected, convolutional, pooling or others.

A bundle $D_i^l$ (\cref{def:bundle}) is then the set of all vectors that are equal accordingly to \cref{eq:bundled}. We are mainly interested in \emph{conflicting} bundles and our aim is to quantify how conflicting a single bundle ${D_i} \in \mathcal{D}^l$ at training step $t$ and layer $l$ is. A standard measure fitted for this task is to evaluate the entropy $H^l(t, D_i)$ as:
\begin{align}
    H^l(t, D_i) = -\sum_{n=1}^{N_c} p(n) \ln \left( p(n) + \epsilon \right)
\end{align}
where $p(n)$ represents the probability that samples of class $n$ occur in bundle $D_i$ and $\epsilon$ is an arbitrarily small value ensuring numerical stability. The value of entropy $H^l(t, D_i)$ is large if the bundle $D_i$ created in layer $l$ contains many examples with different labels. Otherwise, if a small number of examples share the same label, $H^l(t, D_i)$ is accordingly small.

To measure the entropy of all bundles, the size of each bundle must also be considered, because a large conflicting bundle $D_1$ affects the training more than a small conflicting bundle $D_2$. Therefore, we consider the bundle size of each bundle in order to provide a \emph{bundle entropy at training step $t$ for layer $l$} through:
\begin{align}
    H^l(t) = \frac{1}{|\mathcal{B}|} \sum_{D_i\in\mathcal{D}^l} |D_i| \cdot H(t, D_i)
\end{align}
If only one sample or samples of the same class are included in a bundle, $H^l(t)$ is zero. If a fully conflicting bundle occurs then $H^L = \ln(N_c)$.

To be able to check the correctness of \cref{hyp:fully} and \cref{hyp:partially}, we must determine whether conflicts occur during training. Therefore, we evaluate $H^l(t)$ after multiple time steps and report an average value of the bundle entropy that occurred during training. We call this metric the \emph{bundle entropy} $H^l$. Unless otherwise specified, we will use $H^l$ for the bundle entropy of the last hidden layer $L$. It is important to mention that the bundle entropy $H^L > 0$ iff at least two samples with different labels are bundled during training.

\section{Experimental evaluation}\label{sec:experimental}

In \cref{sec:theory} we hypothesized that if conflicts occur, the test loss increases (\cref{hyp:partially}) and in the extreme case all labels would be ignored (\cref{hyp:fully}). In this section, we evaluate whether conflicting bundles occur and their effect during the training of state-of-the-art methods.

\subsection{Setup} \label{sec:experimental_setup}

We believe that the analysis of conflicting bundles can lead to new methods by avoiding the negative effects of conflicting layers. The focus of this paper is therefore not to fine-tune specific state-of-the-art methods but to shed light on this problem. We will consider the following setup for evaluation:

\begin{figure*}[t]
  \centering
  \includegraphics[width=.95\linewidth]{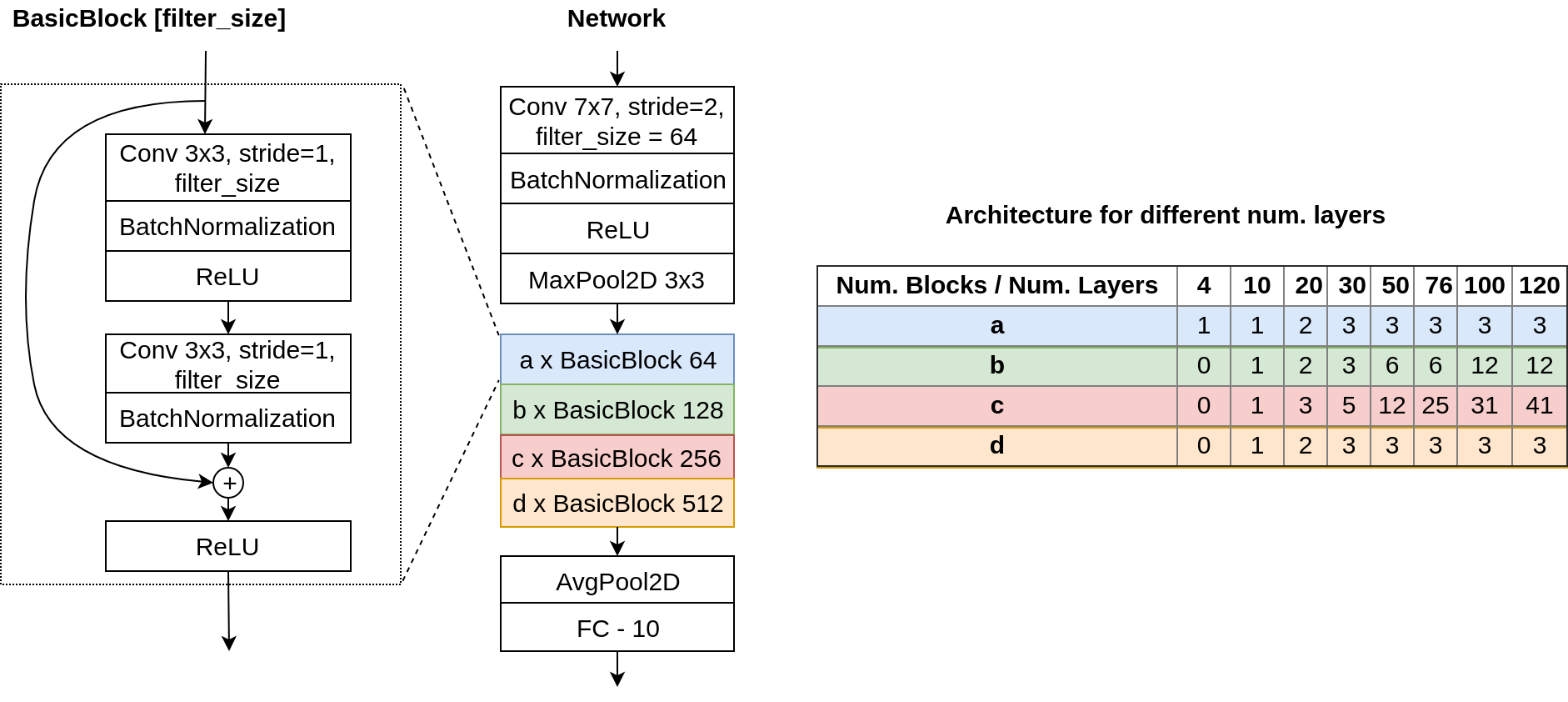}
\caption{ResNet architecture \citet{residual_neural_networks} used in our experimental evaluation.}
\label{fig:architecture}
\end{figure*}

\paragraph{Training.} We will use the term "fully connected networks" - or equivalently "FC-Net" - for networks that are built only with fully connected layers: For networks, such as the ones introduced by \citet{residual_neural_networks} and as shown in \cref{fig:architecture}, the term we will use is "residual neural networks" or "ResNet" in short; and for the same type but without residual connections, "VGG-Net". We will use the most commonly-used datasets \cite{imagenette, cifar10, mnist, svhn} and random data augmentation to evaluate conflicting bundles under a standard setup as follows: We normalize and randomly crop, flip (except for MNIST), and adapt the brightness of images; ReLU activations are used and therefore weights are initialized with the HE initializer \cite{he_initializer} to avoid vanishing or exploding gradients; To minimize the cross-entropy loss we use the state-of-the-art optimizer Ranger (RAdam \cite{radam} + Lookahead \cite{lookahead}) with a mini-batch size of $64$, a learning rate of $0.001$ and weight decay of $0.01$. ResNets and VGG-Nets are trained for $120$ epochs and the FC-Nets for $50$ epochs. The source code is implemented in TensorFlow Version 2.2.0 \cite{tensorflow} and available for download on GitHub\footnote{\url{https://github.com/peerdavid/conflicting-bundles}}. Training is executed on a multi-GPU cluster, where one GPU is used to measure conflicting bundles. All experiments are also implemented and designed to be executable on smaller systems with a single GPU.

\paragraph{Evaluation.} Test accuracy is averaged over the last $5$ epochs in order to exclude outliers. We estimate conflicting bundles through a random subset of the training set $\mathcal{B} \subseteq \mathcal{X}$ with $|\mathcal{B}| = 2048$ to speed up computations and we found empirically that this number of samples is large enough to represent the bundle entropy. To calculate bundles, we created a vectorized function that iterates only once overall $x \in \mathcal{B}$ such that this calculation can also be done on small hardware setups with e.g. only one GPU. Therefore, the complexity to evaluate all layers is $O(|\mathcal{B}| \times L)$. Please note that examples of the training data and not the test data are used to measure conflicting training bundles, as those examples are used to adjust the weights of the network. The developed TensorFlow software module to evaluate conflicting bundles is certified for computational reproducibility\footnote{\url{https://codeocean.com/capsule/8314999/tree/v1}} and is documented in detail in our previous work \cite{conflicting_bundle_module}.

\subsection{Training with a fully conflicting bundle} \label{sec:experimental_fully}

We will first evaluate if training with fully conflicting bundles settles at a region where all neurons fire with a constant value as predicted in \cref{hyp:fully}. For this, we will compare the training of a neural network with and without a fully conflicting bundle under controlled settings. The fully conflicting bundle is produced in a setup where the conditions can be tightly controlled through manual weight initialization. In this first experiment, we use a toy dataset with two classes (class zero if $x_i < 0.5$, class one otherwise) and $1,000$ training examples for a two-layer (two neurons per layer) network with ReLU activation followed by softmax and cross-entropy loss function.

\begin{figure*}[t]
\centering
\begin{subfigure}{.4\textwidth}
  \centering
  \captionsetup{justification=centering}
  \includegraphics[width=.99\linewidth]{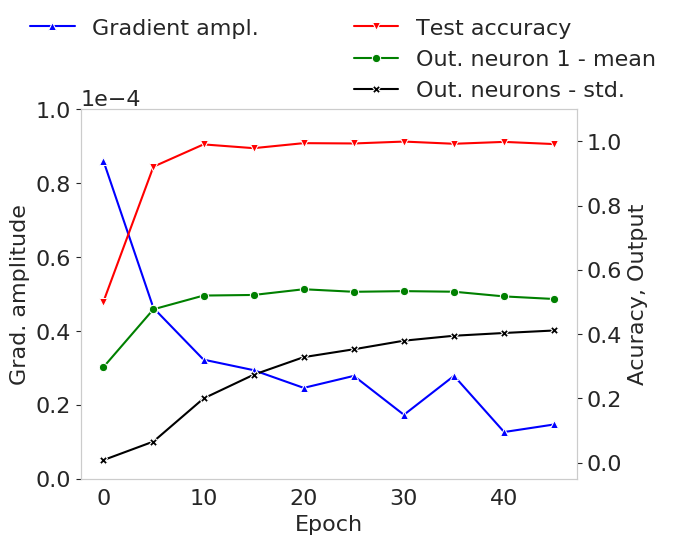}  
  \caption{
  No conflicting bundle \\ balanced dataset.}\label{fig:experiment_threshold_a}
\end{subfigure}
\begin{subfigure}{.4\textwidth}
  \centering
  \captionsetup{justification=centering}
  \includegraphics[width=.99\linewidth]{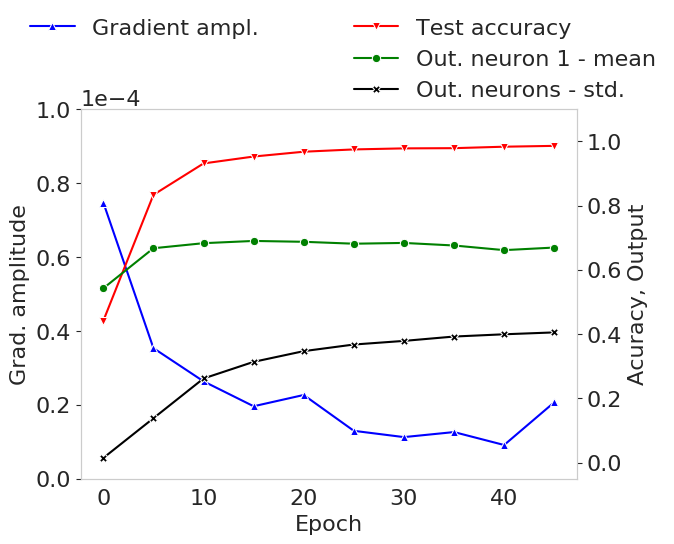}  
  \caption{
  No conflicting bundle \\ imbalanced dataset.}\label{fig:experiment_threshold_b}
\end{subfigure}

\begin{subfigure}{.4\textwidth}
  \centering
  \captionsetup{justification=centering}
  \includegraphics[width=.99\linewidth]{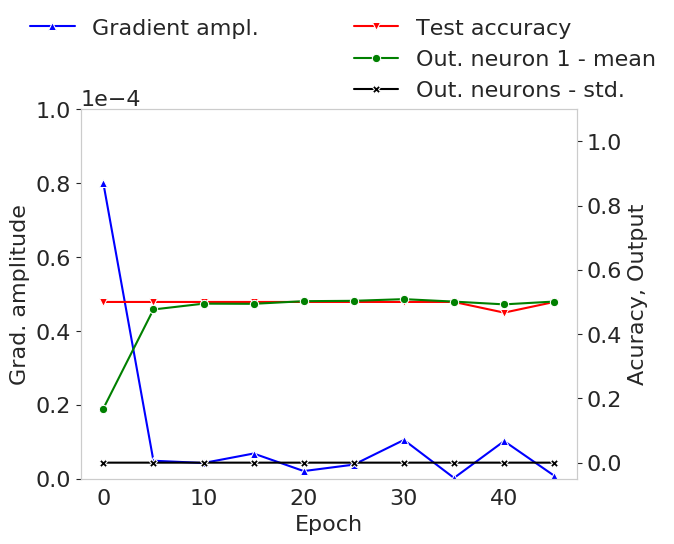}  
  \caption{Fully conflicting bundle \\ training with balanced dataset}
  \label{fig:experiment_threshold_c}
\end{subfigure}
\begin{subfigure}{.4\textwidth}
  \centering
  \captionsetup{justification=centering}
  \includegraphics[width=.99\linewidth]{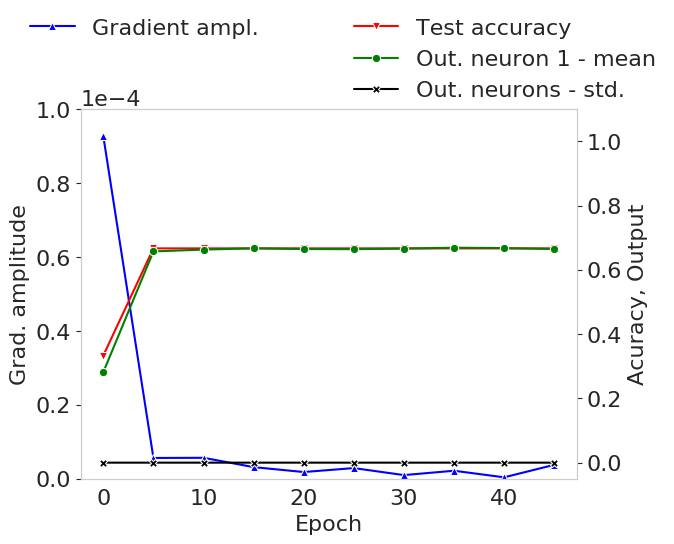}  
  \caption{Fully conflicting bundle \\ training with imbalanced dataset}
  \label{fig:experiment_threshold_d}
\end{subfigure}

\caption{Results using a toy dataset and manually initialized weights with the  absence conflicting bundles for (a) balanced and (b) imbalanced datasets. Compare with (c) and (d) in the presence of a fully conflicing bundle. The gradient (blue line), test accuracy (red line), mean of output neuron 1 (green line) and standard deviation of output neurons (black line) are shown.}
\label{fig:experiment_threshold}
\end{figure*}

As \cref{fig:experiment_threshold_a} shows, this simple network can solve the problem with a training accuracy of $\approx 1.0$ if weights are initialized such that no conflicting bundle is produced. Compare those results with the graph shown in \cref{fig:experiment_threshold_c}, where we initialized weights such that a fully conflicting bundle is produced. It can be seen that for this weight configuration each output neuron fires with a constant value of $1 / N_c = 0.5$ after $50$ epochs as predicted by \cref{hyp:fully}. In \cref{fig:experiment_threshold_b} and \cref{fig:experiment_threshold_d} the network is trained with an imbalanced dataset such that $66\%$ of the training examples are of class zero and $33\%$ are of class one. We can see in \cref{fig:experiment_threshold_b} that each neuron reflects the aforementioned imbalance (\cref{sec:theory_fully}) of the dataset since neuron one fires with a constant value of $0.66$. 

\begin{figure}[t]
  \centering
  \captionsetup{justification=centering}
  \includegraphics[width=.45\linewidth]{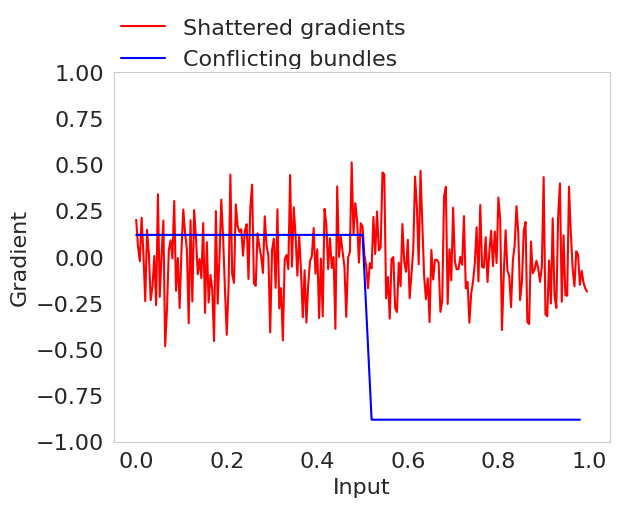}  
  \caption{White noise of shattered gradients \cite{shattered_gradients} (red line) compared with the gradients of the conflicting training bundle problem (blue line).}
  \label{fig:experiment_threshold_shattered}
\end{figure}

The gradient (blue line) of the training without conflicting bundles (fig. \ref{fig:experiment_threshold_a} and \ref{fig:experiment_threshold_b}) and with fully conflicting bundles (fig. \ref{fig:experiment_threshold_c} and \ref{fig:experiment_threshold_d}) is similar at the beginning of the training process, ruling out the gradient vanishing or exploding problem. Also, as networks have only two layers, this problem would by principle not arise. For completeness in excluding we are facing a case of the shattered gradient problem, we further analyzed these results by analyzing white noise in both cases (shattered gradients and conflicting bundles), that is, we evaluated whether inputs and their respective gradients are uncorrelated. If they were, it would indicate that we are facing a case of the shattered gradient problem  \cite{shattered_gradients}. As shown in \cref{fig:experiment_threshold_shattered}, the conflicting bundles and the shattered gradients problems are of different nature since the gradients of conflicting bundles are highly correlated with its input as opposed to the case of shattered gradients.

\subsection{Fully connected networks} \label{sec:experiment_fnn}

\begin{figure*}[t]
\begin{subfigure}{.32\textwidth}
  \centering
  \captionsetup{justification=centering}
  \includegraphics[width=.97\linewidth]{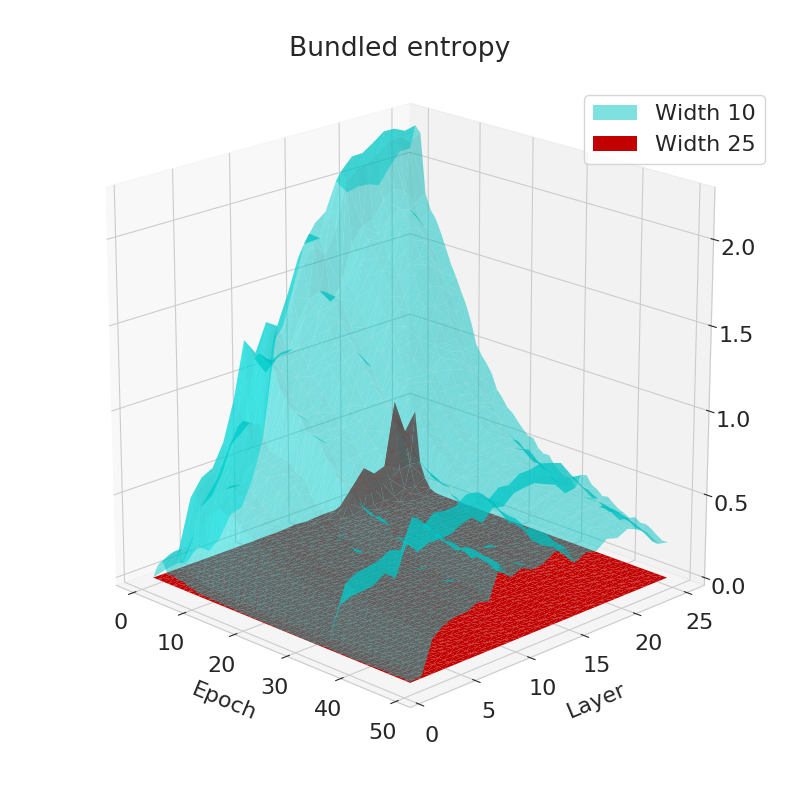}
  \caption{$H^l(t)$ evaluated for each layer $l$ and epoch $t$ for a network with $L=25$.}\label{fig:experiment_fnn_3d_a}
\end{subfigure}
\begin{subfigure}{.32\textwidth}
  \centering
  \captionsetup{justification=centering}
  \includegraphics[width=.97\linewidth]{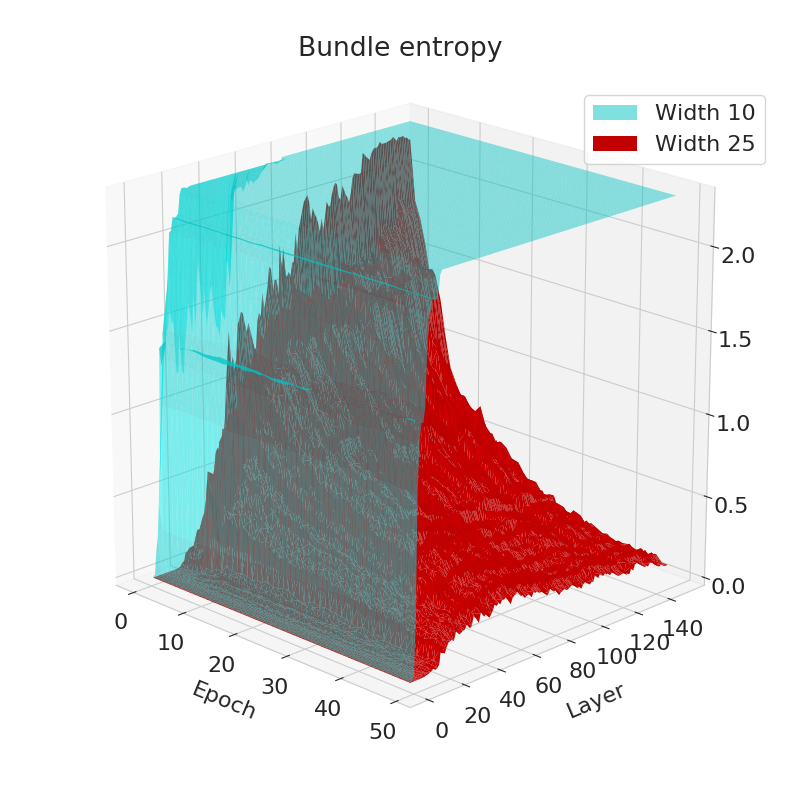}
  \caption{$H^l(t)$ evaluated for each layer $l$ and epoch $t$ for a network with $L=150$.}\label{fig:experiment_fnn_3d_b}
\end{subfigure}
\begin{subfigure}{.32\textwidth}
  \centering
  \captionsetup{justification=centering}
  \includegraphics[width=.97\linewidth]{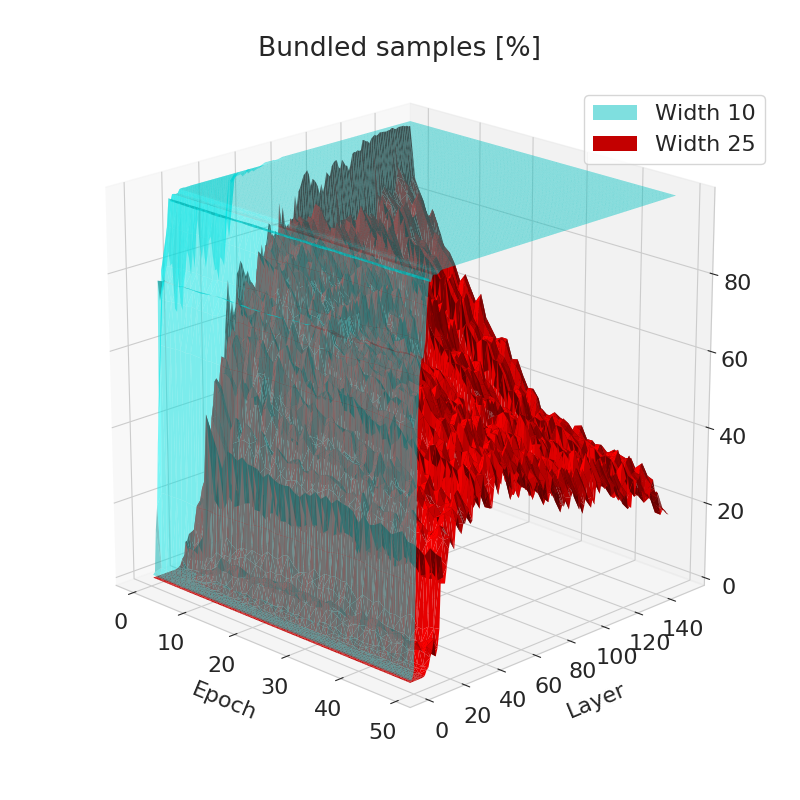}
  \caption{Number of bundles (\cref{def:bundle}) evaluated for each layer $l$ and epoch $t$ for a network with $L=150$.}\label{fig:experiment_fnn_3d_c}
\end{subfigure}
\caption{Evaluation of FC-Nets at different $L$ depths and two widths ($10$ and $25$) w.r.t conflicting training bundles.}
\label{fig:experiment_fnn_3d}
\end{figure*}

\begin{figure}[t]
  \centering
  \captionsetup{justification=centering}
  \includegraphics[width=.5\linewidth]{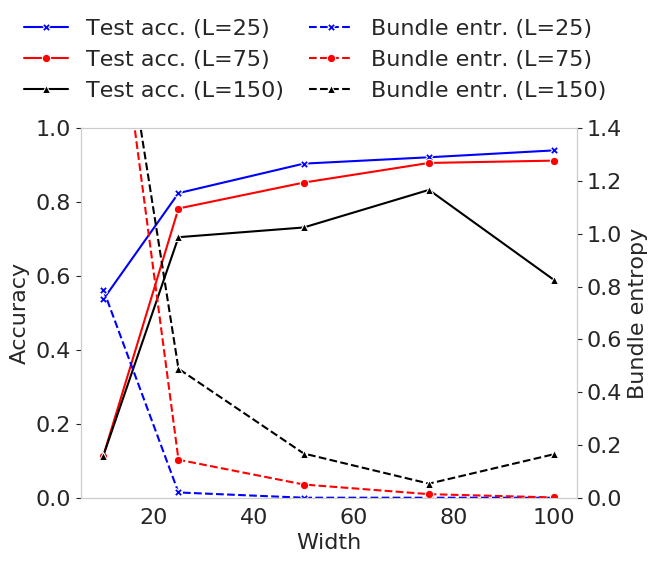}
  \caption{Test accuracy (solid lines) and bundle entropy $H^L$ (dotted lines) for FC-Nets with different depths (L) and widths.}\label{fig:experiment_fnn_width}
\end{figure}

The test accuracy of more than a hundred FC-Nets networks together with the conflicting boundary was already introduced in \cref{fig:experiment_fnn_first_conflicting_layer}. In this section, we study the bundle entropy and the number of bundles that occur during training in more detail by analyzing each training epoch and each layer of different FC-Nets trained on MNIST. Results are shown in \cref{fig:experiment_fnn_3d} and \cref{fig:experiment_fnn_width}.

\Cref{fig:experiment_fnn_3d_a} shows the training of two networks (25 layers) with widths of $10$ and $25$. The number of conflicts increases as we go higher in the hierarchy of the network, indicating that subsequent layers cannot fully solve conflicts. We can then conclude that the number of conflicting training bundles increases as the depth of the network increases. If we are to compare the results of width $10$ with those of width $25$, \cref{fig:experiment_fnn_3d_a} shows that conflicts occur much earlier in the architecture if the dimensionality of hidden features is smaller. This would explain the test-accuracy pattern  shown in \cref{fig:experiment_fnn_first_conflicting_layer}, where the yellow boundary was created - already at the beginning of training - by observing that if conflicting bundles occur during training, conflicts can only be slowly resolved as training proceeds. Two deeper networks with $150$ layers are analyzed in \cref{fig:experiment_fnn_3d_b} and \cref{fig:experiment_fnn_3d_c}. For both widths ($10$ and $25$), conflicting bundles occur during training, but for a width of $10$ it can be seen that a fully conflicting bundle occurs since $H^l(t) \approx 2.3$. Conflicts were never resolved, which was confirmed by the accuracy being not better than chance (\cref{hyp:fully}). Interestingly, if we compare \cref{fig:experiment_fnn_3d_a} and \cref{fig:experiment_fnn_3d_b}, we can conclude that for a fixed width, the position of the layer where conflicting bundles occur is similar among networks with different depths. For example, the first conflicting layer for width $25$ is layer $19$ for both depths, $L=25$ and $L=150$. Therefore, it would be sufficient to only evaluate the deepest network to find the conflicting boundary (\cref{fig:experiment_fnn_first_conflicting_layer}). 

The correlation of the bundle entropy and test accuracy is evaluated in \cref{fig:experiment_fnn_width}, which shows that the bundle entropy $H^L$ is negatively correlated with the test accuracy. This further supports  \cref{hyp:partially}, which stated that conflicting bundles worsen the test accuracy of neural networks. One natural question that arises from the experimental evaluation in this section is whether conflicts also occur for computer vision tasks and convolutional layers because hidden features are high dimensional, which we will analyze next.

\subsection{VGG-Nets}\label{sec:experiment_vgg}

\begin{figure*}[t]
\begin{subfigure}{.48\textwidth}
  \vspace{20pt}
  \centering
  \captionsetup{justification=centering}
  \includegraphics[width=.96\linewidth]{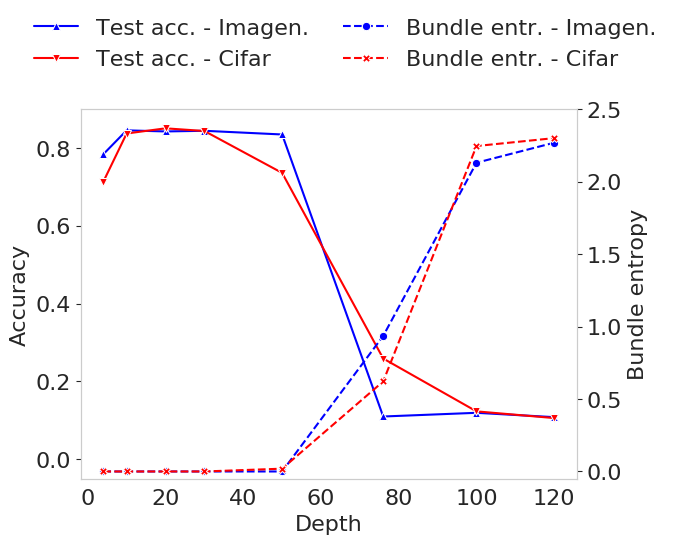}
  \caption{Test accuracy (solid lines) and bundle entropy $H^L$ (dotted lines) for Imagenette (blue lines) and CIFAR (red lines) using different number of layers (Depth) and without residual connections}\label{fig:experiment_vgg_a}
\end{subfigure}
\begin{subfigure}{.48\textwidth}
  \centering
  \captionsetup{justification=centering}
  \includegraphics[width=.9\linewidth]{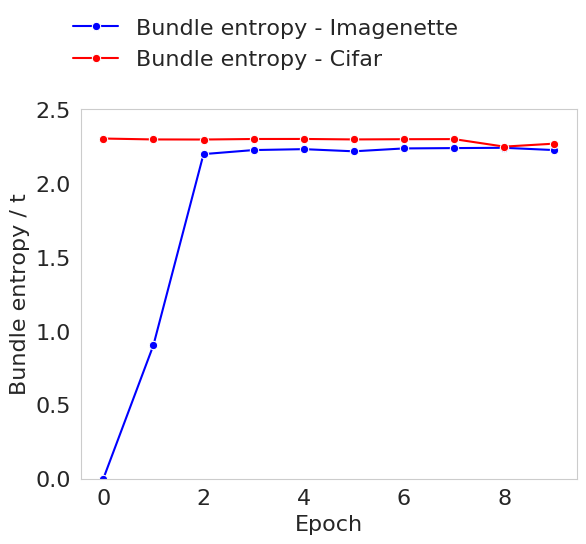}
  \caption{$H^L(t)$ for a $100$ layer VGG Network without \\ residuals trained on CIFAR and Imagenette.}\label{fig:experiment_vgg_b}
\end{subfigure}
\caption{Performance and bundle entropy of VGG-Nets trained on CIFAR and Imagenette.}
\label{fig:experiment_vgg}
\end{figure*}

We evaluate next \cref{hyp:fully} and \cref{hyp:partially} experimentally on several VGG-Nets for different datasets. \Cref{fig:experiment_vgg} shows the performance and bundle entropy for Imagenette and CIFAR (blue and red lines respectively in \cref{fig:experiment_vgg_a}). \Cref{fig:experiment_vgg_b} shows the bundle entropy at each time step $H^L(t)$ for a network with $100$ layers trained on Imagenette.

As \cref{fig:experiment_vgg_a}) shows, for small networks with only four layers, the model suffers from underfitting and therefore the test accuracy is $71\%$ for CIFAR and $78\%$ for Imagenette. From $10$ to $40$ layers, the test accuracy is quite higher and the bundle entropy is zero, which indicates no conflicting bundles during training. After $40$  and $60$ layers for CIFAR and Imagenette respectively, the bundle entropy $H^L$ increases proportional to depth. From this point, the test accuracy decreases proportional to the increase in bundle entropy (\cref{hyp:partially}). For $120$ layers, the entropy is at a peak for $10$ classes, i.e. $\approx 2.3$, which indicates a fully conflicting bundle. In this case, the accuracy is also not better than chance (\cref{hyp:fully}) as shown in \cref{fig:experiment_vgg_a}). Previous work has already reported a drop in test accuracy when training very deep convolutional networks \cite{shattered_gradients, residual_neural_networks, highway_networks}. This behavior can be the result of the appearance of conflicting bundles. If we analyze the progress of training (\cref{fig:experiment_vgg_b}), it can be seen that  conflicting bundles appear after just a few epochs of training. Even though it is well known that a proper weight initialization is critical for a successful training process, these results show that even when weights are initialized correctly, the probability for conflicting layers increases proportionally with each layer added to the architecture (including the case of very high dimensional features such as images). 

It is well known that residual connections reduce the shattered gradient problem. In \cref{sec:experimental_fully} and  \cref{sec:experiment_fnn} we showed that that shattered gradients problem is different from conflicting bundles. We analyze next whether residual connections help also at mitigating the conflicting bundles problem.

\subsection{Residual neural networks}\label{sec:experiment_residual}

\begin{figure}[t]
  \centering
  \captionsetup{justification=centering}
  \includegraphics[width=.55\linewidth]{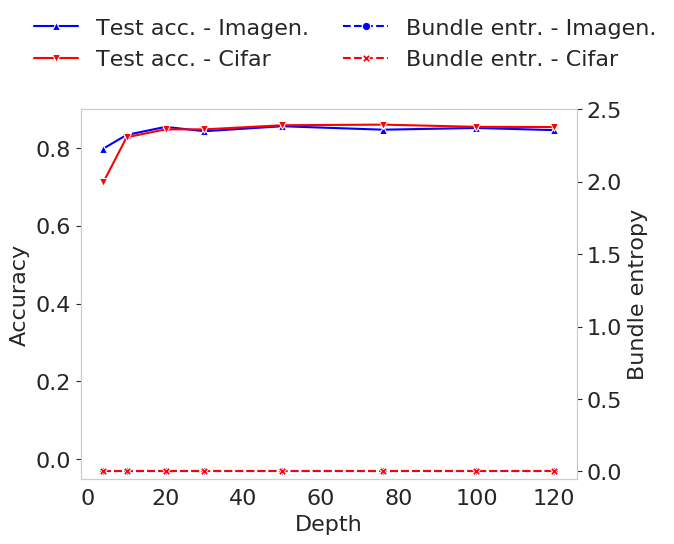}
  \caption{Test accuracy and bundle entropy $H^L$ for different number of layers with residual connections}
  \label{fig:experiment_resnet}
\end{figure}

To evaluate whether residual networks also suffer from conflicting bundles, we trained ResNets with different depths on different datasets. \Cref{fig:experiment_resnet} shows that the bundle entropy is zero for all network depths and datasets, this would indicate that conflicts are solved by residual connections and therefore, the training of very deep convolutional networks is possible with an accuracy similar to all the different residual networks (\cref{hyp:partially}). \citet{residual_neural_networks} already showed that residual networks are easier to optimize, our conflicting bundle analysis provides a new explanation on why that seems to be the case. 

We next evaluated each network block by measuring whether conflicts occur directly after the second batch normalization layer and before the residual connection is added. We found that many blocks ($\approx 60\%$ in a ResNet-120 trained on CIFAR) exist where conflicts are produced directly after the second batch normalization layer. The conflicts seem to be resolved after the residual connection is added to this otherwise conflicting output. For the sake of completeness, we provide an analysis from a theoretical point of view: Let's consider layer $l$ as the one producing conflicts for $x_i$ and $x_j$ in the absence of residual connections. We will call this as the intermediate conflicting output $d^{(l)}$ (which is the same for $x_i$ and $x_j$ as it is assumed to be conflicting). The output for inputs $x_i$ and $x_j$ of the layer which adds a residual connection $r^{(l)}(x)$ is $a^{(l+1)}(x) = r^{(l)}(x) + d^{(l)}$. The residual $r^{(l)}(x)$ is the identity mapping \cite{residual_neural_networks} and therefore it can be shown that the function $a^{(l+1)}(x)$ is bijective for inputs $x_i$ and $x_j$. \Cref{def:bundling} is violated because $a^{(l+1)}_i \neq a^{(l+1)}_j$ and the conflict is resolved as shown empirically.

\begin{figure}[t]
  \centering
  \captionsetup{justification=centering}
  \includegraphics[width=.55\linewidth]{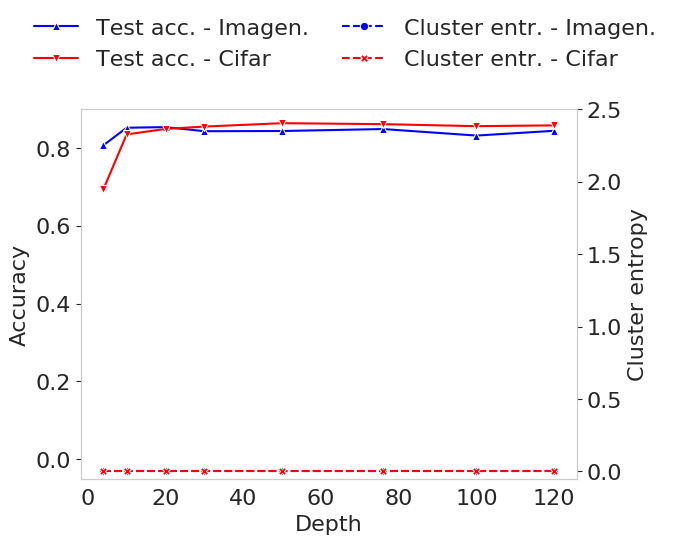}  
  \caption{A network that uses the residual function \\ $r^{(l)}(x) = 2x + 0.1$ } \label{fig:residual_bijective}
\end{figure}

This analysis also holds without the assumption that the residual connection is the identity function as any bijective residual function $r^{(l)}(x)$ is sufficient in order to solve conflicts, which has already been extensively studied by \citet{resnet_identity}. To further confirm this fact, we executed an additional experiment where we trained a residual network on Imagenette with the mapping function $r^l(x) = 2x + 0.1$ to evaluate if a) produces a similar accuracy and, b) it also solves conflicts. \Cref{fig:residual_bijective} shows the results of this experiment. Therefore, our theory extends preliminary work \cite{resnet_norm_preservation} that explicitly states that their theoretical investigation holds only for residuals with identity mappings.  

\begin{figure*}[t]
\begin{subfigure}{.45\textwidth}
  \centering
  \captionsetup{justification=centering}
  \includegraphics[width=.9\linewidth]{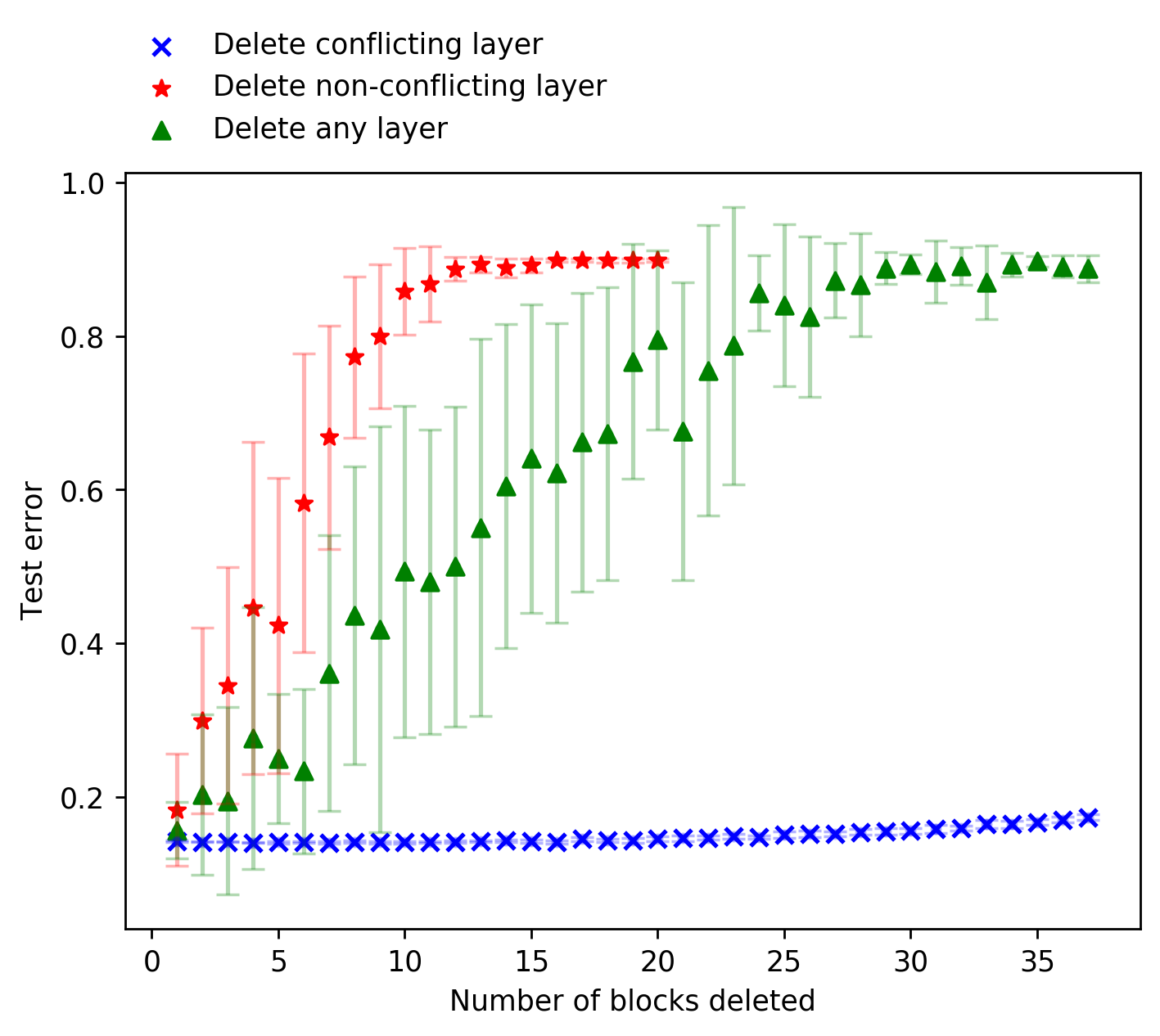}
  \caption{Performance of ResNet-120 trained for $120$ epochs on CIFAR. $n$ blocks (with two layers per block) are deleted of the trained network.}\label{fig:experiment_resnet_lesion_a}
\end{subfigure}
\begin{subfigure}{.45\textwidth}
  \centering
  \captionsetup{justification=centering}
  \includegraphics[width=.9\linewidth]{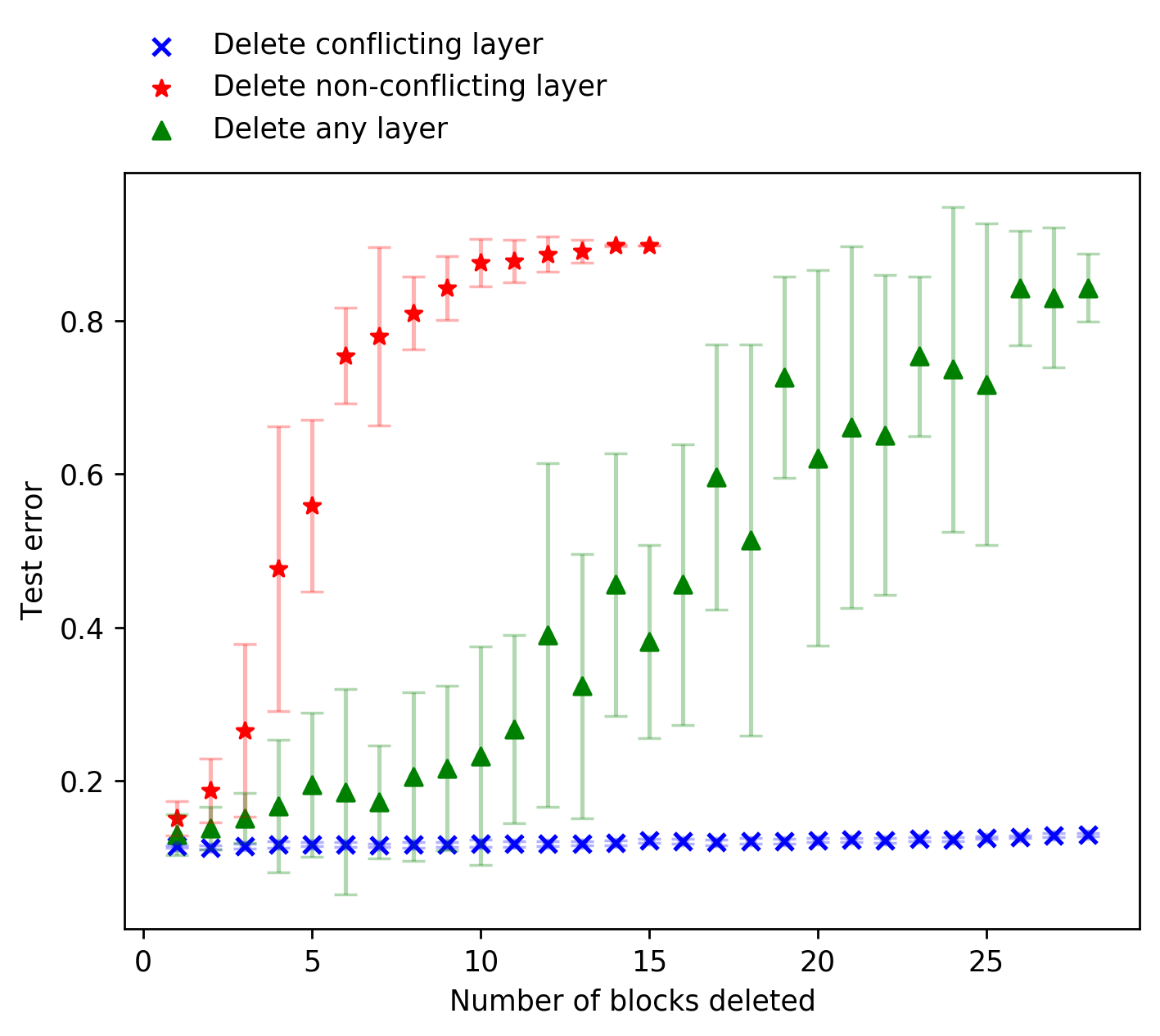}
  
  \caption{Performance of ResNet-120 trained for $120$ epochs on Imagenette. $n$ blocks (with two layers per block) are deleted of the trained network.}\label{fig:experiment_resnet_lesion_b}
\end{subfigure}

\caption{Lesion studies for a trained ResNet-120. We compare lesions of conflicting-layers, non-conflicting layers and any layer following \citet{residual_behave_shallow} and report the mean and the standard deviation of 15 executions.}
\label{fig:experiment_resnet_lesion}
\end{figure*}

Another important fact from this theoretical analysis is, that conflicting layers that are bypassed with residual connections represent only a linear mapping, because $a^{(l+1)}(x) = r^{(l)}(x) + d^{(l)}$. Therefore, it should be possible to completely delete layers that produce conflicts within a residual block from already trained residual networks with only a minor impact on the test-error. \citet{residual_behave_shallow} has already shown that in fact the test-error increases only slightly as more and more layers are deleted from the architecture of trained residual networks. We reproduced those results and additionally evaluate separately how the error increases if (1) we delete conflicting layers or (2) we delete non-conflicting layers. Results of our lesion study are shown in \cref{fig:experiment_resnet_lesion}. It can be seen that more than in fact, $60\%$ of the layers for a Resnet-120 trained on CIFAR and more than $50\%$ when trained on Imagenette can be removed, having a minor effect on the test error (blue crosses in \cref{fig:experiment_resnet_lesion}). This is true only when we delete conflicting and partially-conflicting layers. Otherwise, if non-conflicting layers are deleted from the architecture we can see that the error rises abruptly (red stars in \cref{fig:experiment_resnet_lesion}). On the other hand, if we remove random layers from the architecture (green triangles in \cref{fig:experiment_resnet_lesion}), this rise is not so abrupt but we still observe a large steady rise in the test error. This behavior makes us believe that this latter error increase may not be only due to the fact that ResNets may behave like ensembles of shallow networks \cite{residual_behave_shallow}, but also to that a combination of conflicting and non-conflicting layers are dropped from the architecture (the former having no effect on performance, the latter having a large effect as shown by the red stars). This statement would be further supported by looking at the standard-deviation values, being much larger when random layers are deleted from the architecture when compared to the other two cases (especially when conflicting layers are removed).

Instead of training those networks and pruning them after the training, it would be of much more interest for the sake of efficiency to have at our disposal a pruning algorithm that removes those conflicting layers already at the beginning of the training.

\subsection{Auto-tuning the depth of a network}

The \emph{conflicting bundle auto-tune} (CBA-tune) pruning algorithm that we will present here looks first for the conflicting boundary - e.g. the yellow border from   \cref{fig:experiment_fnn_first_conflicting_layer} - for a given input dimensionality. Afterwards, it will iteratively prune the network to ensure that no conflicting bundles occur during training. The pseudocode for CBA-tune is shown in \cref{alg:cba} and works as follows: First, the requested network would be trained for at least one epoch, e.g. the one from \cref{sec:experiment_vgg} with $120$ layers. Then, the first conflicting layer $l$ with  $H^l(t) > 0$ and all subsequent layers of the same block type (as specified in \cite{residual_neural_networks}) are removed from the architecture (\cref{sec:experiment_fnn}). A side effect of this pruning is that also the dimensionality between two layers changes and therefore, we need to restart the training with the new pruned architecture rather than continuing the training to avoid dimensionality problems. We then re-initialize all weights with the HE initializer \cite{he_initializer}, because the network is trained for one epoch with conflicting bundles such that weights of the network are adjusted into wrong directions. This process is repeated until no conflicting layer can be found and the network is successfully trained for 120 epochs.

\begin{algorithm*}[t] 
	\caption{CBA-tune} \label{alg:cba}
	\begin{algorithmic}[1]
	\State {model $\gets$ VGG-Net with $a=3$, $b=12$, $c=41$, $d=3$} \Comment{Initially create the largerst VGG-Net with $120$ layers.}
	\State {Initialize weights of model} \Comment{The HE-Initializer \cite{he_initializer} is used to initialize weights.}
	\State {$i \gets 0$}
	\State { } 
	\While {$i < 120$}
	    \State {Train model for one epoch}
	    \State {$i \gets i+1$}
	    \State { } 
        \If {\ Conflicting bundles occurred during training}
            \State {$l \gets$ First conflicting layer w.r.t block-type} \Comment{$0 < l < 120$}
            \State {$t \gets$ Block-type of layer $l$} \Comment{$t \in \{a,b,c,d\}$}
            \State {Prune block-type $t$ of the current model to size $l$} \Comment{Prune only one block-type per iteration.}
            \State {Initialize weights of model}
            \State{$i \gets 0$}
        \EndIf
	\EndWhile
	\end{algorithmic} 
\end{algorithm*}

\begin{table*}[t]
  \caption{Comparison of test accuracy, memory consumption (based on checkpoint size) and inference time.}
  \small
  \label{tbl:auto_tune}
  \centering
  \begin{tabular}{llc|ccc}
    \hline
    Dataset & Name     & Layers     & Accuracy [\%] & Mem. [MB] & Time / Step [ms] \\
    \hline
    Imagenette & ResNet & $50$ & $85.6 \pm 0.8$ & $116$ & $310$    \\
               & \textbf{Auto-tune} & $\pmb{24 \pm 1.6}$ & $\pmb{84.7 \pm 0.1}$ & $\pmb{63 \pm 6.0}$ & $\pmb{200 \pm 0.0}$ \\
    \hline
    Cifar & ResNet & $76$ & $85.7 \pm 0.4$ & $178$ & $130$   \\
          & \textbf{Auto-tune} & $\pmb{19 \pm 0.9}$ & $\pmb{85.3 \pm 0.4}$ & $\pmb{53 \pm 1.0}$ & $\pmb{48 \pm 2.5}$ \\
    \hline
    Svhn & ResNet & $50$ & $95.4 \pm 0.1$ & $116$ & $90$    \\
               & \textbf{Auto-tune} & $\pmb{19 \pm 0.9}$ & $\pmb{95.1 \pm 0.1}$ & $\pmb{54 \pm 0.5}$ & $\pmb{46 \pm 1.8}$    \\
    \hline
    Mnist & ResNet & $50$ & $ 99.4 \pm 0.0$ & $116$ & $88$ \\
               & \textbf{Auto-tune} & $\pmb{ 17 \pm 1 }$ & $\pmb{ 99.3 \pm 0.5 }$  & $\pmb{54 \pm 2.1}$ & $\pmb{ 40 \pm 0.0}$    \\
    \hline
  \end{tabular}
\end{table*}

As an example, we will explain here the process followed for the network used in  \cref{sec:experiment_vgg} with $120$ layers. In order to show the validity of our CBA-tune approach, we will compare the pruned network found by the auto-tune algorithm with the ResNet that produced the highest accuracy (ResNet-$50$ as in \cref{sec:experiment_residual}). In addition to the CIFAR and Imagenette datasets, we evaluated the SVHN \cite{svhn} and MNIST \cite{mnist} datasets for completeness. Each experiment is run three times and we report the mean and standard deviation. \Cref{tbl:auto_tune} shows that the different architectures found by our CBA-tune algorithm differ only slightly between different runs as shown in \cref{tbl:auto_tune_multiple_executions}. Our CBA-tune algorithm changed the architecture at most three times before the final pruned network was found and the architecture was changed after the first and before the second epoch. As an example, the evolution for one execution of CBA-tune algorithm trained on CIFAR is shown in \cref{tbl:auto_tune_evolution}. Our automatic depth selection process is computationally very efficient as it took only three epochs of training to find the conflict-free architecture. The number of layers found by the CBA-tune algorithm is within the optimal region w.r.t. test accuracy (\cref{fig:experiment_vgg_a}) and the depth increases proportional to its input dimensionality (\cref{sec:experiment_fnn}). The number of layers obtained by CBA-tune also corresponds to the ones presented by \citet{residual_behave_shallow}, that is,  paths in ResNets are just between 10 to 34 layers deep. \Cref{tbl:auto_tune} further shows that the test accuracy of the networks with conflicting layers removed is not only comparable to their residual counterpart, but also the inference time and the memory consumption is drastically reduced since fewer layers are used. 

\begin{table}[t]
  \centering
  \small
  \caption{Evolution of the architecture as created by CBA-tune shown for Cifar. The numbers show how many basic blocks (as shown in \cref{fig:architecture}, left side) are still used after the pruning was executed.}
  \label{tbl:auto_tune_evolution}
  \begin{tabular}{l|cccc}
    \hline
    Epoch & Block a & Block b & Block c & Block d \\
    \hline
    Start & 3 & 12 & 41 & 3 \\
    1 & 3 & 4 & 41 & 3 \\
    2 & 3 & 3 & 41 & 3 \\
    \textbf{3} & \textbf{3} & \textbf{3} & \textbf{0}
    & \textbf{3} \\
    \hline
  \end{tabular}
\end{table}

\begin{table}[t]
  \centering
  \small
  \caption{Three different executions of CBA-tune together with the corresponding architecture that was used for training.}
  \label{tbl:auto_tune_multiple_executions}
  \begin{tabular}{l|ccccccccc}
    \hline
    & & & \multicolumn{4}{c}{Block} & & &                  \\
    Dataset & Run & Layers & a & b & c & d & Accuracy [\%] & Mem. [MB] & Time [ms] \\
    \hline
    Imagenette & 1 & 26 & 3 & 3 & 3 & 3 & 84.5 & 69 & 200 \\
               & 2 & 22 & 3 & 4 & 0 & 3 & 84.7 & 55 & 200 \\
               & 3 & 24 & 3 & 3 & 2 & 3 & 84.8 & 65 & 200 \\
    \hline
    Cifar & 1 & 20 & 3 & 3 & 0 & 3 & 85.3 & 54 & 50 \\
          & 2 & 18 & 3 & 2 & 0 & 3 & 84.9 & 53 & 45 \\
          & 3 & 20 & 3 & 3 & 0 & 3 & 85.8 & 53 & 50 \\
    \hline
    Svhn & 1 & 20 & 3 & 3 & 0 & 3 & 95.3 & 54 & 47 \\
         & 2 & 20 & 3 & 3 & 0 & 3 & 95.1 & 54 & 47 \\
         & 3 & 18 & 3 & 2 & 0 & 3 & 95.0 & 53 & 43 \\
    \hline
    Mnist & 1 & 18 & 3 & 1 & 1 & 3 & 99.2 & 57 & 40 \\
          & 2 & 16 & 3 & 1 & 0 & 3 & 99.3 & 52 & 40 \\
          & 3 & 16 & 3 & 1 & 0 & 3 & 99.3 & 52 & 40 \\
    \hline
  \end{tabular}
\end{table}

\section{Discussion and future work}\label{sec:discussion}

In this paper, we defined and introduced the problem of \emph{conflicting training bundles}. In \cref{sec:theory} this problem is analyzed theoretically and we hypothesized that (1) conflicting bundles that occur during training decrease the accuracy of trained models (\cref{hyp:partially}) and (2) fully conflicting bundles can lead to networks that cannot be trained at all (\cref{hyp:fully}). In the experimental \cref{sec:experimental} both hypotheses were evaluated empirically on many different hyperparameters, architectures, and datasets. We first defined a toy dataset under controlled settings to be able to analyze the behavior of neural networks. Later, we  showed empirically for Fully connected networks trained on MNIST that the number of conflicting training bundles increases proportional to (1) the depth of the network and (2) a decreasing  dimensionality of hidden features. We have also shown that conflicts occur early in the training and that the bundle entropy is negatively correlated with the accuracy of the trained model. Our findings showed that conflicting bundles do not shatter gradients, or in other words, the shattering gradients problem \cite{shattered_gradients} is different from the conflicting bundles problem presented here. VGG-Nets were then evaluated in \cref{sec:experiment_vgg} to confirm that conflicting-bundles can also occur when hidden features are very high-dimensional, as it is the case for images. Conflicting layers can be bypassed with residual connections, and we proved that any bijective residual connection can be used, not only the identity function  \cite{residual_neural_networks, resnet_identity, resnet_norm_preservation}. Conflicting layers that are bypassed with residuals produce a linear mapping between its input and its output. We could easily remove those layers from already trained networks, such that as much as $60\%$ of the layers of a residual network could be removed without increasing the accuracy which is an alternative explanation to the idea that residual networks behave like ensembles of shallow networks \cite{residual_behave_shallow}. The CBA-tune we have presented in this work automatically prunes VGG-Nets to avoid conflicting layers as early as at the beginning of the training. With this computationally efficient algorithm, the accuracy of the network models was maintained while the computational power and memory consumption was drastically reduced.

The findings and insights of this paper can be of great help for future AutoML methods. For example, mutated architectures can be rejected after a few training steps in evolutionary algorithms if conflicting bundles occur following \cref{hyp:partially}. Additionally,  architectures can be precisely adapted because the layer(s) causing conflict(s) could be known. We believe that the analysis of conflicting bundles will help future researchers to design better deep neural network architectures. In this paper, we pruned the depth of the network to avoid conflicting layers. In future work, it would be interesting to also explore alongside other dimensions of the hyperparameter space such as the learning-rate, the batch-size, or the width of the network. The presented work has shown that residuals force a linear mapping between its input and its outputs and therefore those layers can also be removed as we demonstrated in our lesion experiment, which could lead to future studies that could transform conflicting layers into layers not containing conflicts.


\section*{Acknowledgments}
We acknowledge all members of the IIS research group, the European Union’s Horizon 2020 program for the grant agreement no. 731761 (IMAGINE) and DeepOpinion for the opportunity to continue with this research in the future.

\bibliography{main}

\end{document}


\title{Conflicting Bundles: Adapting Architectures Towards the Improved  \\ Training of Deep Neural Networks - Supplementary Material}

\author{David Peer \qquad
Sebastian Stabinger \qquad
Antonio Rodr\'{i}guez-S\'{a}nchez\\
University of Innsbruck \\
Austria\\
{\tt\small https://iis.uibk.ac.at/}
}

\maketitle



\section*{2.2 \ \ Conflicting bundle metric}

The following approximation for definition 1 is motivated in the paper to check if two samples $x_i$ and $x_j$ are bundled:
\begin{align} \label{eq:metric}
    \frac{\alpha}{|\mathcal{B}|} \ ||a_i^{(l+1)} - a_j^{(l+1)}||_\infty \leq \gamma
\end{align}
To further support this claim, we demonstrate next using a heatmap how many bundles can be correctly detected if the finite representation is considered or if the finite representation is neglected. In a second experiment we compare \cref{eq:metric} that considers the resolution during backpropagation with the check $a_i^{(l+1)} = a_j^{(l+1)}$ that not considers this.

\begin{figure}[H]
\begin{subfigure}{.33\textwidth}
  \centering
  \captionsetup{justification=centering}
  \includegraphics[width=.99\linewidth]{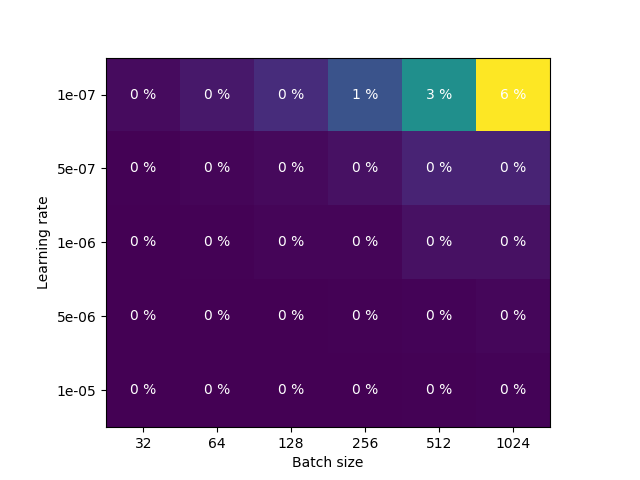}  
  \caption{(Generated) real conflicts }\label{fig:conflicts_a}
\end{subfigure}
\begin{subfigure}{.33\textwidth}
  \centering
  \captionsetup{justification=centering}
  \includegraphics[width=.99\linewidth]{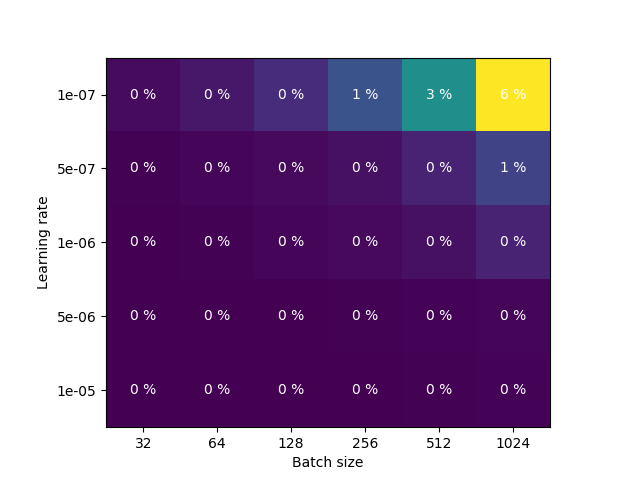}  
  \caption{Bundles measured with \cref{eq:metric} to consider the resolution during backpropagation}
  \label{fig:conflicts_b}
\end{subfigure}
\begin{subfigure}{.33\textwidth}
  \centering
  \captionsetup{justification=centering}
  \includegraphics[width=.99\linewidth]{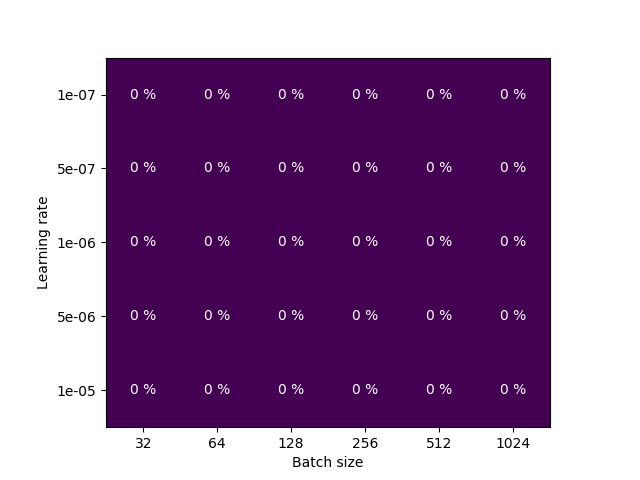}  
  \caption{Bundles measured with $a^{(L)}_i = a^{(L)}_j$ which not consideres the resolution during backpropagation}
  \label{fig:conflicts_c}
\end{subfigure}
\caption{A heatmap that shows how many bundles are detected with different metrics.}
\label{fig:conflicts}
\end{figure}

\paragraph{Forward vs. backpropagation}
We sampled $60$k random $a^{(L)}$ values from $[0, 1)$ and randomly created weights within $[0.5, 1.0)$. For different learning rates and batch sizes, if there are two $a^{(L)}_i$ and $a^{(L)}_j$ values that update the weights $W$ exactly the same way, we report them as bundled in \cref{fig:conflicts_a}. Afterwards, we report how many bundles can be detected by simply using the $a_i^{(l+1)} = a_j^{(l+1)}$ check in \cref{fig:conflicts_c}. We report the same for the proposed method \cref{eq:metric} in \cref{fig:conflicts_b}. 
It can be seen in \cref{fig:conflicts_b} that the floating-point representation must be considered to be able to measure the real formation of bundles during backpropagation, otherwise bundles are not detected as shown in \cref{fig:conflicts_c}.

\begin{figure}[H]
\begin{subfigure}{.45\textwidth}
  \centering
  \captionsetup{justification=centering}
  \includegraphics[width=.7\linewidth]{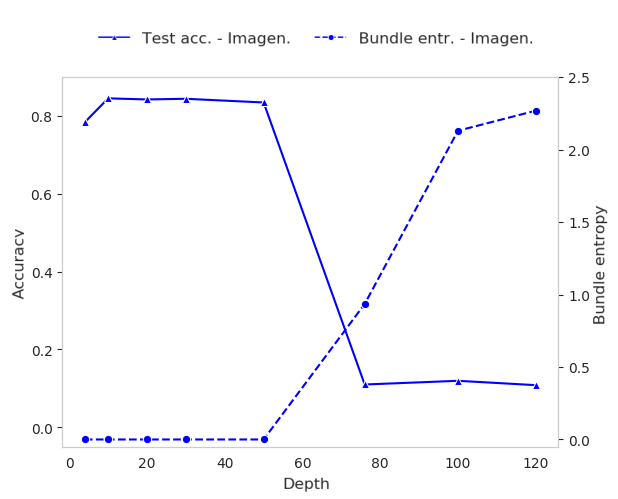}  
  \caption{Bundle entropy measured with \cref{eq:metric}.}\label{fig:experiment_bundling_real_a}
\end{subfigure}
\begin{subfigure}{.45\textwidth}
  \centering
  \captionsetup{justification=centering}
  \includegraphics[width=.7\linewidth]{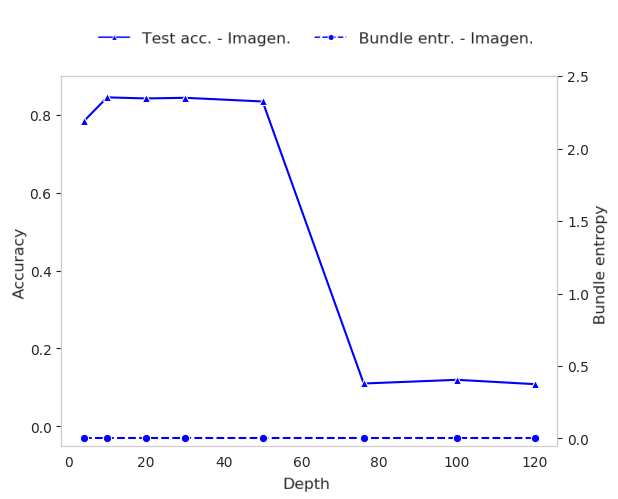}  
  \caption{Bundle entropy measured with $a^{(L)}_i = a^{(L)}_j$ }\label{fig:experiment_bundling_real_b}
\end{subfigure}
\caption{Comparison of the bundle entropy if the finite representation of the used floating points value is considered or not.}
\label{fig:experiment_bundling_real}
\end{figure}

\paragraph{Measure the bundle entropy} In this experiment, we compare the bundle entropy that is measured using  $a^{(L)}_i = a^{(L)}_j$ with the bundle entropy measured using \cref{eq:metric}. The result is shown in \cref{fig:experiment_bundling_real}.

It can be seen that bundles are only detected if the real representation that is used in CPUs and GPUs is considered also for this real-world example. Otherwise, bundling happens during backpropagation, but it cannot be quantified correctly.

\section*{3.1 \ \ Setup}

\begin{figure}[H]
  \centering
  \includegraphics[width=.8\linewidth]{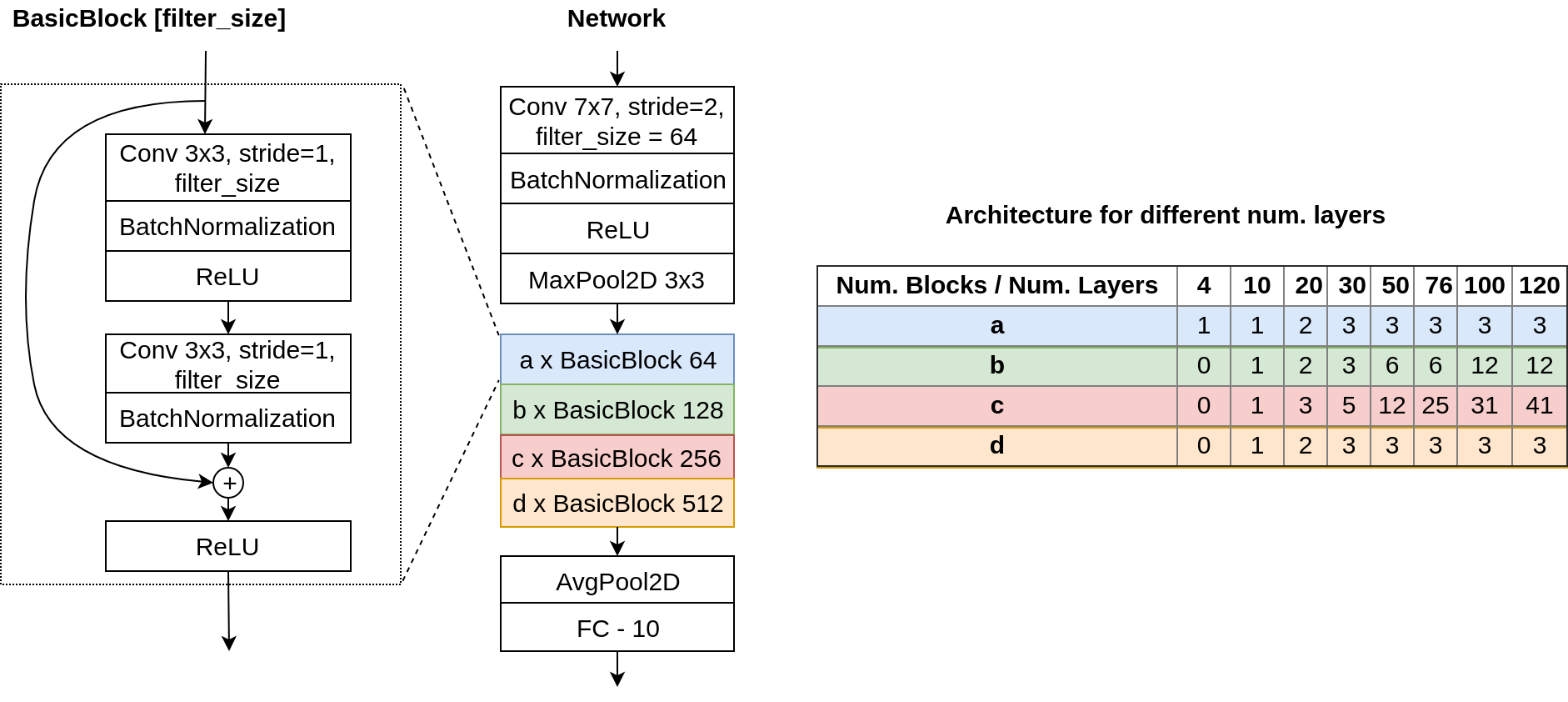}
\caption{Architecture motivated by \citet{residual_neural_networks} that is used for experimental evaluation.}
\label{fig:architecture}
\end{figure}

In image \cref{fig:architecture} the architecture that is used for VGG net (without residual connection) and ResNet is shown. We followed \citet{residual_neural_networks} to design the basic blocks and scale the number of layers.

\section*{3.2 \ \ Training with a fully conflicting bundle}
\paragraph{Detailed results.} In the following graphs we show detailed training results of each experiment that we executed for the fully conflicting bundle case under controlled settings:

\begin{figure}[H]
\centering
\begin{subfigure}{.4\textwidth}
  \centering
  \includegraphics[width=.9\linewidth]{img/experiments/threshold/learning_True_learning.png}  
  \caption{No conflicting bundle, balanced dataset.}\label{fig:experiment_threshold_a}
\end{subfigure}
\begin{subfigure}{.4\textwidth}
  \centering
  \includegraphics[width=.9\linewidth]{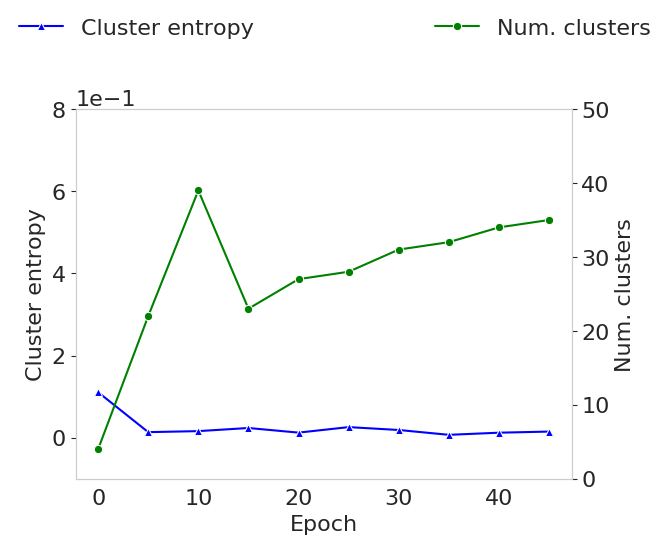}  
  \caption{No conflicting bundle, balanced dataset.}
  \label{fig:experiment_threshold_b}
\end{subfigure}

\begin{subfigure}{.4\textwidth}
  \centering
  \includegraphics[width=.9\linewidth]{img/experiments/threshold/learning_False_learning.png}  
  \caption{No conflicting bundle, inbalanced dataset.}\label{fig:experiment_threshold_c}
\end{subfigure}
\begin{subfigure}{.4\textwidth}
  \centering
  \includegraphics[width=.9\linewidth]{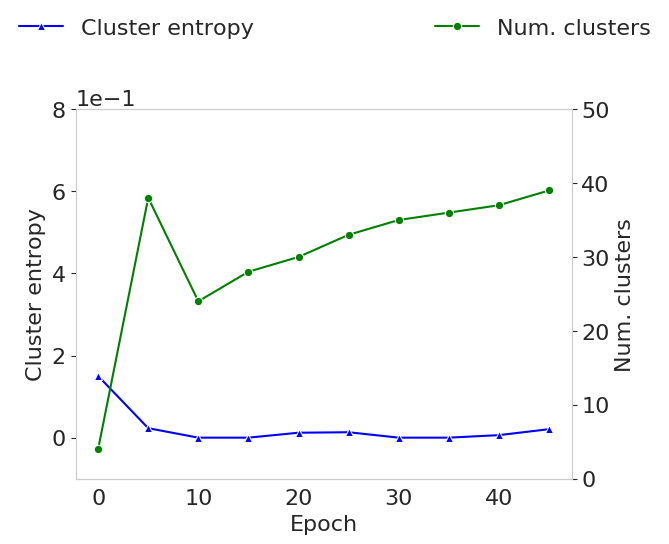}  
  \caption{No conflicting bundle, inbalanced dataset.}
  \label{fig:experiment_threshold_d}
\end{subfigure}

\begin{subfigure}{.4\textwidth}
  \centering
  \includegraphics[width=.9\linewidth]{img/experiments/threshold/conflicts_True_learning.png}  
  \caption{Fully conflicting bundle, balanced dataset.}\label{fig:experiment_threshold_e}
\end{subfigure}
\begin{subfigure}{.4\textwidth}
  \centering
  \includegraphics[width=.9\linewidth]{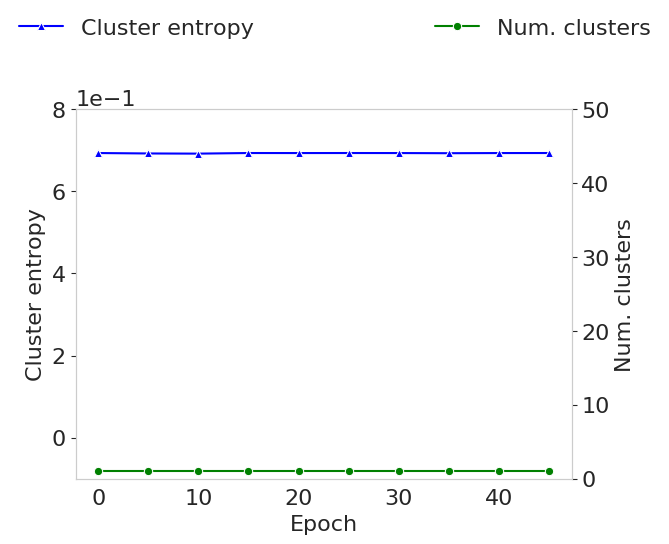}  
  \caption{Fully conflicting bundle, balanced dataset.}
  \label{fig:experiment_threshold_f}
\end{subfigure}

\begin{subfigure}{.4\textwidth}
  \centering
  \includegraphics[width=.9\linewidth]{img/experiments/threshold/conflicts_False_learning.png}  
  \caption{Fully conflicting bundle, balanced dataset.}\label{fig:experiment_threshold_g}
\end{subfigure}
\begin{subfigure}{.4\textwidth}
  \centering
  \includegraphics[width=.9\linewidth]{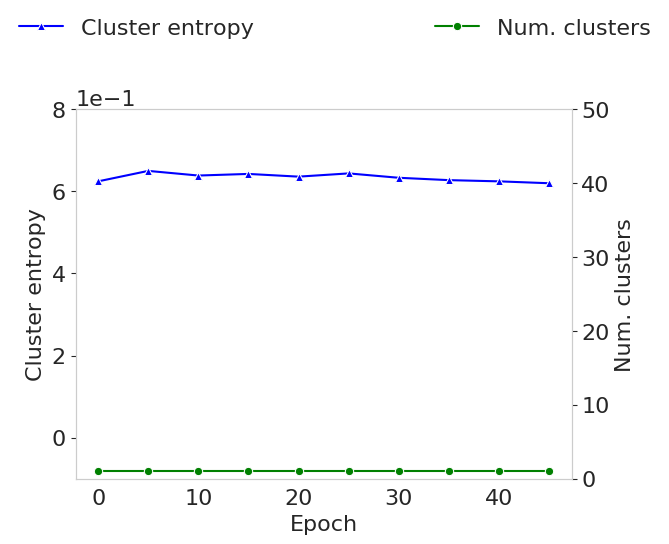}  
  \caption{Fully conflicting bundle, balanced dataset.}
  \label{fig:experiment_threshold_h}
\end{subfigure}

\caption{Experiment with a toy dataset and manually initialized weights to compare training with a fully conflicting bundle for balanced and imbalanced datasets.}
\label{fig:experiment_threshold}
\end{figure}

\paragraph{Weight initialization.}
For this experiment, it is important to ensure that a single fully conflicting bundle is created in order to be able to evaluate if the predictions of our hypothesis are correct. The weights are initialized as follows:

\begin{figure}[H]
  \centering
  \includegraphics[width=.5\linewidth]{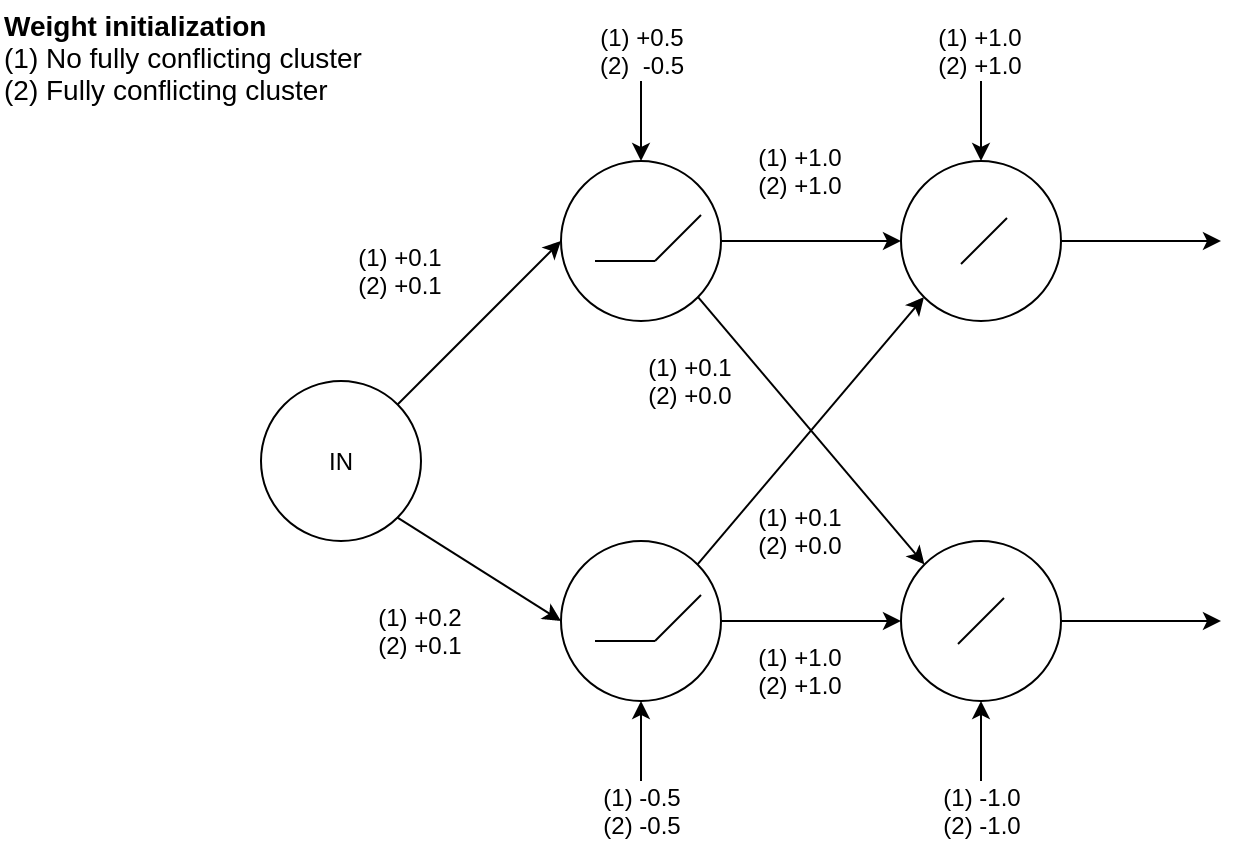}
\caption{Initialization of the weights to not produce conflicting bundles (1) or to produce a fully conflicting bundle (2)}
\label{fig:weight_init}
\end{figure}

\section*{3.5 \ \ Residual connections}

\begin{figure}[H]
  \centering
  \includegraphics[width=.5\linewidth]{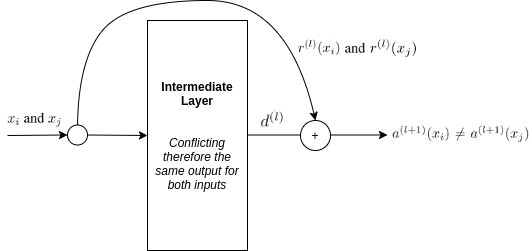}
\caption{Graphical illustration of the proof that bijective residual functions bypass conflicting layers.}
\label{fig:resnet_proof}
\end{figure}

\paragraph{Detailed proof that residuals bypass conflicts.}
We prove that residual connections bypass or solve conflicting layers. A graphical idea of the proof is given in \cref{fig:resnet_proof}. 
Additionally, we relax the assumption that the bypass is an identity mapping as this strong assumption is not needed for this proof, we simply assume that the bypass or residual mapping $r(x)^l$ for layer $l$ is some bijective function:

\begin{proof}
Assume that layer $l$ produces conflicts for $x_i$ and $x_j$ without a residual connection. We call this intermediate conflicting output $d^{(l)}$. The output for $x_i$ and $x_j$ of the layer which adds a residual connection $r^{(l)}(x)$ is therefore $a^{(l+1)}(x) = r^{(l)}(x) + d^{(l)}$. The residual $r^{(l)}(x)$ is a bijective function \cite{residual_neural_networks} and therefore it can be shown that the output $a^{(l+1)}(x)$ is bijective for inputs $x_i$ and $x_j$:

\emph{Injective:} $a^{(l+1)}(x)$ is injective for inputs $x_i$ and $x_j$ iff $a^{(l+1)}(x_i) = a^{(l+1)}(x_j) \rightarrow x_i = x_j$. We know that $r^{(l)}(x_i) + d^{(l)} = r^{(l)}(x_j) + d^{(l)}$ iff $r^{(l)}(x_i)  = r^{(l)}(x_j)$ iff $x_i = x_j$, because $r^{(l)}(x)$ is assumed to be bijective. Therefore $a^{(l+1)}(x)$ is injective on inputs $x_i$ and $x_j$.

\emph{Surjective: } $a^{(l+1)}(x)$ is surjective for inputs $x_i$ and $x_j$ iff $\forall y \ \exists x. \ a(x) = y$.
But for all $y$ there exists some $x$ such that $y - d^{(l)} = r^{(l)}(x)$ since $r^{(l)}(x)$ is assumed to be bijective. Therefore $a^{(l+1)}(x)$ is injective on inputs $x_i$ and $x_j$.

\emph{Bijective: }
$a^{(l+1)}(x)$ is bijective for inputs $x_i$ and $x_j$ because it is surjective and injective for those inputs. Therefore $a^{(l+1)}(x_i) \neq a^{(l+1)}(x_j)$ which conflicts the definition of bundles and it can be concluded that conflicts are solved (or bypassed) if the residual $r^{(l)}(x)$ is a bijective mapping.
\end{proof}

We proved that \emph{conflicts introduced by a layer are solved (or bypassed) if a bijective residual connection is added}. Note that we \textbf{not} proved that \emph{conflicts are impossible if bijective residual connections are added}, although the experiments indicate that this happens in practice. More precisely, it is not possible to prove this stronger statement because there exists a counterexample: If the learned intermediate result is exactly the negative function of the residual, the sum of the intermediate result and the residual is always zero and therefore fully conflicting, but this case is very unlikely and therefore, we have never seen this case in the experiments  (see also \citet{residual_neural_networks}) .

\section*{3.6 \ \ Auto-tune}

\paragraph{Evolution of the architecture.} 
In this table, we give one detailed execution, how the auto-tune algorithm pruned the $120$ layer network architecture. The table below shows the adaption for the network trained on Cifar (Execution $1$).
\begin{table}[H]
  \centering
  \begin{tabular}{l|cccc}
    \hline
    Epoch & Block a & Block b & Block c & Block d \\
    \hline
    Start & 3 & 12 & 41 & 3 \\
    1 & 3 & 4 & 41 & 3 \\
    2 & 3 & 3 & 41 & 3 \\
    \textbf{3} & \textbf{3} & \textbf{3} & \textbf{0}
    & \textbf{3} \\
    \hline
  \end{tabular}
\end{table}
It can be seen that only three epochs are needed in order to find this architecture, although the new architecture is faster, needs less memory and maintains the accuracy as shown in the paper.

\paragraph{Multiple executions.}
In the tables below we report each architecture that the auto-tune algorithm created for each dataset and three different executions:

Imagenette:
\begin{table}[H]
  \label{tbl:auto_tune_imagenette_multi}
  \centering
  \begin{tabular}{ccccccccc}
    \hline
    & & \multicolumn{4}{c}{Block} & & &                 \\
    Execution & Layers & a & b & c & d & Accuracy & Mem. [MB] & Time / Step [ms] \\
    \hline
    1 & 26 & 3 & 3 & 3 & 3 & 84.5 & 69 & 200 \\
    2 & 22 & 3 & 4 & 0 & 3 & 84.7 & 55 & 200 \\
    3 & 24 & 3 & 3 & 2 & 3 & 84.8 & 65 & 200 \\
    \hline
  \end{tabular}
\end{table}

Cifar:
\begin{table}[H]
  \label{tbl:auto_tune_cifar_multi}
  \centering
  \begin{tabular}{ccccccccc}
    \hline
    & & \multicolumn{4}{c}{Block} & & &                  \\
    Execution & Layers & a & b & c & d & Accuracy [\%] & Mem. [MB] & Time / Step [ms] \\
    \hline
    1 & 20 & 3 & 3 & 0 & 3 & 85.3 & 54 & 50 \\
    2 & 18 & 3 & 2 & 0 & 3 & 84.9 & 53 & 45 \\
    3 & 20 & 3 & 3 & 0 & 3 & 85.8 & 53 & 50 \\
    \hline
  \end{tabular}
\end{table}

Svhn:
\begin{table}[H]
  \label{tbl:auto_tune_imagenette_multi}
  \centering
  \begin{tabular}{ccccccccc}
    \hline
    & & \multicolumn{4}{c}{Block} & & &                 \\
    Execution & Layers & a & b & c & d & Accuracy & Mem. [MB] & Time / Step [ms] \\
    \hline
    1 & 20 & 3 & 3 & 0 & 3 & 95.3 & 54 & 47 \\
    2 & 18 & 3 & 2 & 0 & 3 & 95.0 & 53 & 43 \\
    3 & 20 & 3 & 3 & 0 & 3 & 95.1 & 54 & 47 \\
    \hline
  \end{tabular}
\end{table}

Mnist:
\begin{table}[H]
  \label{tbl:auto_tune_imagenette_multi}
  \centering
  \begin{tabular}{ccccccccc}
    \hline
    & & \multicolumn{4}{c}{Block} & & &                 \\
    Execution & Layers & a & b & c & d & Accuracy & Mem. [MB] & Time / Step [ms] \\
    \hline
    1 & 18 & 3 & 1 & 1 & 3 & 99.2 & 57 & 40 \\
    2 & 16 & 3 & 1 & 0 & 3 & 99.3 & 52 & 40 \\
    3 & 16 & 3 & 1 & 0 & 3 & 99.3 & 52 & 40 \\
    \hline
  \end{tabular}
\end{table}

{\small
\bibliographystyle{plainnat}
\bibliography{egbib}
}